%% file: main.tex
\pdfoutput=1

\documentclass[10pt,twocolumn,letterpaper]{article}

\usepackage[pagenumbers]{wacv} %

\usepackage[hidelinks]{hyperref}

\input{preamble}

\definecolor{wacvblue}{rgb}{0.21,0.49,0.74}

\def\wacvPaperID{2580} %
\def\confName{WACV}
\def\confYear{2027}

\title{SCCV-6D: Sparse Canonical Coordinate Voting for Holistic Multi-View RGB-D 6D Pose Estimation in Robotic Bin Picking}

\author{Nico Leuze, Samed Doğan, Maximilian Hoh, Henry Schaub, Alfred Schoettl\\
Institute for Applications of Machine Learning and Intelligent Systems\\
University of Applied Science Munich, 80335 Munich, Germany\\
{\tt\small nico.leuze@hm.edu, alfred.schoettl@hm.edu}
}

\begin{document}
\maketitle
\input{sec/0_abstract}    
\input{sec/1_intro}

\input{sec/2_relatedworks}
\input{sec/3_method}
\input{sec/4_results_experiments}
{
    \small
    \bibliographystyle{ieeenat_fullname}
    \bibliography{main}
}

\end{document}

%% file: preamble.tex
\usepackage{amsmath}
\usepackage{amssymb}
\usepackage{booktabs}
\usepackage{multirow}
\usepackage{array}
\usepackage{graphicx}
\usepackage{tikz}
\usepackage{pgfplots}
\usepackage[T1]{fontenc}
\usepackage{times}

\newcommand{\methodname}{SCCV-6D}
\newcommand{\prior}{SDT-6D}

\newcommand{\R}{\mathbb{R}}
\newcommand{\SO}{\mathrm{SO}(3)}
\newcommand{\SE}{\mathrm{SE}(3)}
\newcommand{\bp}{\mathbf{p}}
\newcommand{\bq}{\mathbf{q}}
\newcommand{\br}{\mathbf{r}}
\newcommand{\bt}{\mathbf{t}}
\newcommand{\bc}{\mathbf{c}}
\newcommand{\bv}{\mathbf{v}}
\newcommand{\bR}{\mathbf{R}}
\newcommand{\bH}{\mathbf{H}}
\newcommand{\bU}{\mathbf{U}}
\newcommand{\bV}{\mathbf{V}}
\newcommand{\bS}{\mathbf{S}}
\newcommand{\bff}{\mathbf{f}}

\newcommand{\bu}{\mathbf{u}}
\newcommand{\bm}{\mathbf{m}}

\newcommand{\bW}{\mathbf{W}}
\newcommand{\bP}{\mathbf{P}}

%% file: sec/0_abstract.tex
\begin{abstract}
Industrial bin picking requires pose estimates that are accurate under clutter, partial occlusion, and the reflective, texture-less nature of manufactured parts. Most recent approaches to 6D pose estimation follow multi-stage pipelines. These achieve high accuracy, but incur memory and runtime footprints that do not cope with the operational constraints of many real-world industrial deployments. Holistic pipelines estimate poses in a single pass, with cost controlled through indiscriminate subsampling or lowered resolution. Continuing this line, we present SCCV-6D, a fully sparse multi-view RGB-D estimator. We reconstruct the 3D space at millimeter precision, prune background via a staged heatmap mechanism, and predict class, translation offset and rotation at highest resolution for an arbitrary number of objects simultaneously. Rotation is formulated as canonical coordinate voting. Points vote for their canonical object coordinates with an associated confidence, enabling closed‑form pose recovery with a symmetry‑aware weighted Procrustes solve. Along an in‑cluster point recovery induced by offset votes, this removes time-consuming post‑processing stages. We reuse the task-supervised attention priors from the heatmap formulation for a novel attention-gated distillation objective that reweights teacher–student feature alignment. On the IPD test split we report competitive $0.826$ AP at $1.13$\,s per scene, significantly faster than the leading entries. On MV-YCB our approach reaches $96.1$ ADD-S, and remains competitive even on the densely packed bins of ROBI. 
\end{abstract}

%% file: sec/1_intro.tex
\section{Introduction}
\label{sec:intro}

\noindent The 6D pose estimation problem in bin picking is subject to  the simultaneous presence of heavy occlusion, high inter-class similarity among industrial parts, and challenging attributes such as specularity and symmetry. Single-view observations frequently capture only fragmented evidence, so pose estimates fall short of the sub-centimeter, sub-degree accuracy required by downstream tasks such as robotic grasping \cite{hodan2018bop, yang2021robi, kalra2026threept, henry_iros}. 

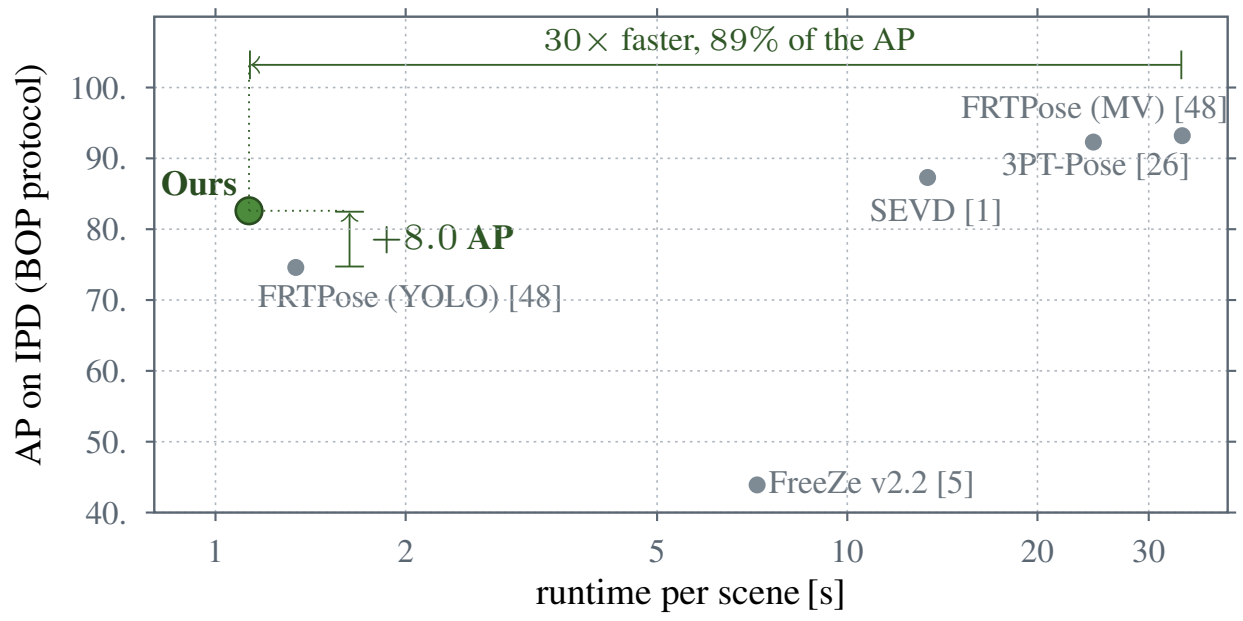
\begin{figure}[t]
  \centering
  \input{sec/tikz_figures/fig1}
  \caption{\textbf{Fast \emph{and} accurate.} SCCV-6D estimates every pose in a single forward pass and a batched closed-form solve. It reaches $0.826$~AP on the IPD test split in $1.13$~s per scene, $30\times$ faster than the leading entry at $89\%$ of its accuracy. SCCV-6D leads the only competitor in the same latency band by $8.0$~AP. Runtimes of competing entries are self-reported on heterogeneous hardware and indicate a trend rather than a matched comparison.}
  \label{fig:teaser}
\end{figure}

\noindent These challenges motivate multi-view acquisition. Surfaces missing in one view are regularly recovered in another, and fusing several calibrated views produces a more consistent and reliable reconstruction \cite{duffhauss2023symfm6d, labbe2020cosypose, duffhauss2022mv6d, yang2021robi, leuze2026sdt6d}. Yet, especially multi-view reconstruction introduces a computational bottleneck. At resolutions sufficient for reliable instance separation, reconstructing and processing the 3D volume incurs substantial computational and memory costs. The reconstruction volume covers the entire workspace while discriminative geometry is localized in small areas, so the task-relevant occupancy is highly sparse. Two paradigms, multi-stage and holistic pipelines, have emerged to handle high-resolution reconstruction differently. 

\noindent Multi-stage pipelines obtain object hypotheses from an upstream, decoupled detector or segmenter, estimate the 6D pose per crop, and append refinement stages \cite{caraffa2024freeze, kalra2026threept, labbe2020cosypose, labbe2022megapose, lin2024sam6d, gao2021cloudaae}. These methods currently lead the industrial track of the BOP challenge \cite{nguyen2025bop} and achieve the highest reported accuracies. However, they are ensembles of independently trained networks. Memory and training-data requirements accumulate across the sub-modules. Inference proceeds sequentially over candidates and views, so runtime scales with the number of objects, views and with the resolution of the crops. \Cref{fig:teaser} plots accuracy against runtime for the leading entries on IPD~\cite{kalra2024ipd}. The two are coupled. No published entry above $0.80$~AP runs in under $13.4$~s, and the fastest entry outperforms the most accurate by $18.6$~AP. Runtimes are self-reported under differing hardware, modalities and view counts, so the entries can not be compared directly. The trend over the population is nonetheless consistent, and it repeats on the other BOP-industrial dataset, XYZ-IBD~\cite{xyz-ibd} and ITODD-MV~\cite{mvitodd}. We attribute the coupling to the sequential structure. Accuracy in such pipelines is gained by appending stages, and the cost is incurred anew for every hypothesis. The fast and accurate regime is unoccupied.

\noindent Holistic pipelines instead estimate all poses in a single forward pass \cite{duffhauss2023symfm6d, breyer2021vgn,duffhauss2022mv6d, he2021ffb6d, he2020pvn3d, li2023swindepose, leuze2026sdt6d}. These methods are considerably smaller and faster, where the computation and memory demand is controlled by indiscriminate scene‑reduction strategies to a fixed cardinality \cite{duffhauss2023symfm6d, duffhauss2022mv6d, li2023swindepose, he2021ffb6d} or coarse‑resolution processing \cite{breyer2021vgn}. Accuracy degrades accordingly \cite{nguyen2025bop}. 

\noindent The authors of \cite{leuze2026sdt6d} proposed to make this reduction step task-dependent via a staged heatmap formulation. A coarse voxelized region-of-interest stage predicts where target objects are plausible, a finer objectness stage focuses towards foreground candidates. Only the retained candidates are forwarded to the 6D pose heads. Allocating computation on task‑relevant evidence rather than indiscriminate sampling strategies appears to be a promising way out of the trade-off. We adopt this line of research and address one structural limitation. \cite{leuze2026sdt6d} consumes depth-only, and geometrically degenerate parts are reported as dominant failure mode. We fuse color and depth directly during scene reconstruction. Synchronized RGB-D views are integrated into a sparse truncated signed distance field \cite{zhou2018open3d}, and the colored surface points from the zero-crossing band are the input to our network. The staged heatmap strategy is applied on the sparse voxelization of that cloud, after which foreground voxel candidates are lifted back to point level through an exact grid lookup. A shared Point Transformer trunk \cite{wu2024ptv3} feeds three heads: classification, translation and rotation. To the best of our knowledge, we present the first fully sparse holistic RGB-D framework for 6D pose estimation.

\noindent Conventional holistic designs recover the pose in a subsequent object-wise aggregation, commonly a mean-shift clustering of keypoint votes followed by a least-squares fit \cite{he2020pvn3d,he2021ffb6d,duffhauss2022mv6d,duffhauss2023symfm6d,li2023swindepose}. Depth-based pipelines frequently append an ICP refinement that dominates the post-processing budget \cite{gao2021cloudaae, leuze2026sdt6d}. We remove this time-consuming stage in three steps. First, translation offsets are voted per point, which naturally forms clusters \cite{leuze2026sdt6d}. A single clustering pass over the scene replaces the per-object mode. Only points voting into a cluster are retained, which sharpens the point cloud segments beyond the heatmap accuracy. Second, every point of a cluster predicts its location in the class-conditioned canonical frame together with a learned confidence. Thus, a point and its predicted coordinate form a 3D-3D correspondence. Third, the pose follows in closed form as the confidence-weighted Procrustes solution \cite{horn1987closed} over the correspondence set that is over-determined by orders of magnitude relative to keypoint voting, batched over all objects of a scene. Assignment and reliability are both predicted rather than inferred from the current pose estimate, so neither correspondence search nor outlier rejection is iterated. 

\noindent Additionally, we introduce heatmap‑guided feature distillation, a novel scheme that improves efficiency while reducing memory consumption. The staged heatmap formulation provides a sparse, task-supervised measure of relevance. We reweight the teacher-student feature alignment by the attention priors assigned by the teacher. The teacher and student share grid construction and network components but differ in capacity. 
\noindent We summarize our contributions as follows.
\begin{itemize}
  \setlength{\itemsep}{1pt}
  \item A holistic, fully sparse multi-view RGB-D framework for 6D pose estimation that fuses appearance and geometry before the first network stage, and recovers poses scene-adaptively in $1.13$~s, outperforming methods in the same latency regime by a wide margin.
  \item A symmetry-aware canonical coordinate voting head that replaces rotation regression by per-point correspondence votes with learned confidences and a weighted Procrustes solve, yielding poses for an arbitrary number of objects in closed form and without iterative refinement.
  \item A heatmap-guided feature distillation scheme that reuses the attention priors of the pipeline as transfer weights.
\end{itemize}

%% file: sec/tikz_figures/fig1.tex
\definecolor{sccvSlate}{HTML}{5A6672}
\definecolor{sccvMist} {HTML}{AEB6BE}
\definecolor{sccvGrey} {HTML}{7E8A94}
\definecolor{sccvGreen}{HTML}{4B8B3B}
\providecommand{\sccvlbl}[1]{{\fontsize{6.2}{7}\selectfont #1}}
\usetikzlibrary{patterns.meta}

\begin{tikzpicture}
\begin{axis}[
  scale only axis, width=7.10cm, height=3.28cm, at={(1.17cm,0.92cm)},
  xmode=log, log basis x=10,
  xmin=0.8, xmax=40, ymin=0.40, ymax=1.10,
  xlabel={runtime per scene\,[s]},
  ylabel={AP on IPD (BOP protocol)},
  xtick={1,2,5,10,20,30}, xticklabels={1,2,5,10,20,30},
  ytick={0.4,0.5,0.6,0.7,0.8,0.9,1.0},
  yticklabels={40.,50.,60.,70.,80.,90.,100.},
  axis line style  = {draw=sccvSlate, line width=0.45pt},
  tick style       = {draw=sccvSlate, line width=0.4pt},
  grid style       = {draw=sccvMist, line width=0.3pt,
                      dash pattern=on 0.5pt off 0.9pt},
  tick label style = {font=\fontsize{6.2}{7}\selectfont, text=sccvSlate},
  label style      = {font=\fontsize{7}{8}\selectfont, text=black},
  ylabel style     = {yshift=-12pt}, xlabel style={yshift=6pt},
  tick align=outside, tickwidth=1.6pt, axis on top,
  ymajorgrids, xmajorgrids, clip mode=individual,
]

  \addplot[only marks, mark=*, mark size=1.4pt, draw=sccvGrey, fill=sccvGrey,
           forget plot]
    coordinates {(24.52,0.923) (33.9,0.932) (13.4,0.873) (7.20,0.439) (1.34,0.746)};

  \addplot[only marks, mark=*, mark size=2.5pt, draw=sccvGreen!55!black,
           fill=sccvGreen, line width=0.5pt, forget plot]
    coordinates {(1.13,0.826)};

  \node[anchor=south west, text=sccvGreen!55!black, inner sep=2.4pt,
        font=\fontsize{6.6}{7.6}\selectfont\bfseries]
        at (axis cs:0.78,0.828) {Ours};
  \node[anchor=north west, text=sccvGrey!85!black, inner sep=2pt]
        at (axis cs:1.12,0.738) {\sccvlbl{FRTPose (YOLO)~\cite{wang2025frtpose}}};
  \node[anchor=west, text=sccvGrey!85!black, inner sep=2pt]
        at (axis cs:7.20,0.439) {\sccvlbl{FreeZe v2.2~\cite{caraffa2024freeze}}};
  \node[anchor=north west, text=sccvGrey!85!black, inner sep=2pt]
        at (axis cs:10.4,0.859) {\sccvlbl{SEVD~\cite{sevd2025}}};
  \node[anchor=south east, text=sccvGrey!85!black, inner sep=2pt]
        at (axis cs:41.85,0.932) {\sccvlbl{FRTPose (MV)~\cite{wang2025frtpose}}};
  \node[anchor=south east, text=sccvGrey!85!black, inner sep=2pt]
        at (axis cs:36.4,0.853) {\sccvlbl{3PT-Pose~\cite{kalra2026threept}}};

  \draw[draw=sccvGreen!70!black, line width=0.35pt,
        |<-|]
        (axis cs:1.13,1.032) -- (axis cs:33.9,1.032);
  \node[anchor=south, text=sccvGreen!62!black, inner sep=1.2pt]
        at (axis cs:6.5,1.036) {\sccvlbl{$30\times$ faster, $89\%$ of the AP}};

  \node[anchor=west, text=sccvGreen!62!black, inner sep=1.6pt,
        font=\fontsize{6.6}{7.6}\selectfont\bfseries]
        at (axis cs:1.7,0.786) {$+8.0$ AP};

    \draw[draw=sccvGreen!70!black, line width=0.35pt,
    |->|]
    (axis cs:1.63,0.746) -- (axis cs:1.63,0.826);

    \draw[draw=sccvGreen!70!black, line width=0.35pt, densely dotted]
    (axis cs:1.13,0.826) -- (axis cs:1.63,0.826);
    \draw[draw=sccvGreen!70!black, line width=0.35pt, densely dotted]
    (axis cs:1.13,0.826) -- (axis cs:1.13,1.032);
    
\end{axis}
\end{tikzpicture}

%% file: sec/2_relatedworks.tex
\section{Related Work}
\label{sec:related}

\paragraph{Industrial 6D pose estimation.} 6D pose estimation in industrial contexts generally falls into two categories, distinguished by how objects are localized. Multi-stage pipelines rely on image or point cloud crops from an upstream detector. CosyPose~\cite{labbe2020cosypose} detects per view, estimates a pose per detection and reconciles the candidates by bundle adjustment, while~\cite{labbe2022megapose, wen2024foundationpose} render and compare against normalized crops. SAM-6D~\cite{lin2024sam6d} and FreeZe~\cite{caraffa2024freeze} pair promptable segmentation or frozen foundation descriptors with a registration step. AlignPose~\cite{alignpose2025} chains a detector, a segmenter, a single-view estimator and two refiners before multi-view aggregation. 3PT~\cite{kalra2026threept} unifies detection, segmentation and pose in a jointly trained transformer pair that tops eleven BOP tracks. Yet, all of these architectures incur structural training and memory
overheads from multiple networks. Sequential candidate processing makes runtime scale with object and view count, and refinement crop size. Holistic pipelines avoid the upstream dependency and estimate all poses in one forward pass. PVN3D~\cite{he2020pvn3d} and FFB6D~\cite{he2021ffb6d} vote for object keypoints from fused appearance and geometry and solve a least-squares fit per object.  MV6D~\cite{ duffhauss2022mv6d} and SyMFM6D~\cite{duffhauss2023symfm6d} extend point-wise voting to several RGB-D views with learned point-to-pixel fusion and, in the latter case, symmetry-aware keypoint targets. SwinDePose~\cite{li2023swindepose} follows the same scheme from depth-only. The respective fusion modules expect an input of fixed cardinality, which forces a random or farthest-point reduction. Methods that consume cropped segments \cite{gao2021cloudaae, shi2021stablepose, lin2024transpose} avoid this by an instance-level assumption. We address this with an approach that keeps the single-pass property of the second line while adopting a task-centric reduction strategy~\cite{leuze2026sdt6d}. 

\paragraph{Sparse encoding for 3D perception.} Sparse processing is the de-facto standard in long-range perception. Approaches differ mainly in how they enlarge the receptive field without densifying, by attending inside cubic windows~\cite{fan2022sst,lai2022stratified}, by partitioning space along the sensor geometry~\cite{lai2023sphereformer}, or by removing dense stages from detection~\cite{chen2023voxelnext}. Point-based operators are sparse by construction. Yet, previous models were dominated by the cost of neighborhood queries~\cite{qi2017pointnet,ptv2}. Point Transformer V3~\cite{wu2024ptv3} replaces those queries by serializing points along space-filling curves, so that a patch is a contiguous slice of a sorted array rather than a gathered neighborhood, and its attention runs over a fixed token count per patch. The receptive field is therefore set and controllable by the grid size and grows through pooling at constant attention cost. The model takes point sets of varying cardinality without padding and scales with the retained points rather than the workspace volume, which is what a scene-adaptive pipeline requires and why we adopt PTv3 as the encoding blocks of the pose heads.

\paragraph{Correspondence-based 6D pose recovery.} A network can regress the parameters of $\SE$ directly. Rotations have no smooth low-dimensional parametrization, so nearby orientations can map to distant target values, and symmetry leaves the target multimodal~\cite{zhou2019continuity}. Correspondence-based predictors delegate pose recovery to a geometric solver, which keeps the learning problem local. That route has been developed primarily on the image lattice, where object coordinates are regressed per pixel~\cite{brachmann2014learning,park2019pix2pose,wang2021gdrnet}. NOCS normalizes the object frame across a category before a closed-form alignment~\cite{wang2019nocs}. Symmetry is resolved once per class, since a crop is not associated with an individual object before the solve. We transfer the idea to the 3D domain, where our translation offset votes group points into object-centric segments that approximate the canonical model up to a rigid transform. Predicting the canonical coordinate of each point makes the correspondence 3D to 3D. The residual is then a Euclidean distance in the metric frame of the reconstruction. The solver is the closed-form weighted Procrustes solution~\cite{kabsch1976,horn1987closed,choy2020dgr}, and the symmetry group is selected per detected object.

\paragraph{Knowledge distillation for point clouds.} Knowledge distillation uses a pretrained teacher to guide a smaller student and has proven effective for training compact models with minimal accuracy loss~\cite{hinton2015distilling, zhang2023pointdistiller, yang2022towards, hou2022pvkd}. Yet, distillation on point clouds remains structurally underexplored. The unordered, sparse nature prevents index‑aligned feature matching, and background dominates the active sites, so a uniformly averaged feature loss is spent almost entirely on regions that carry no task-relevant information. The limited local structure of a single point further weakens the supervision a feature‑matching loss can transfer. Existing schemes therefore guide the transfer. Yang~\etal~\cite{yang2022towards} restrict the loss to positions selected by the teacher's classification response, PVKD~\cite{hou2022pvkd} distills point-to-voxel relations for LiDAR segmentation, and PointDistiller~\cite{zhang2023pointdistiller} combines graph-based local distillation with an occupancy-weighting strategy. We believe that this proxy does not carry over to bin picking. Densely reconstructed regions, e.g. bin walls, produce higher occupancy weights than the small, reflective and frequently occluded target objects. We weight instead by the task-supervised, object-centric attention priors of the heatmap stages, available at no additional cost.

%% file: sec/3_method.tex
\section{Method}
\label{sec:method}

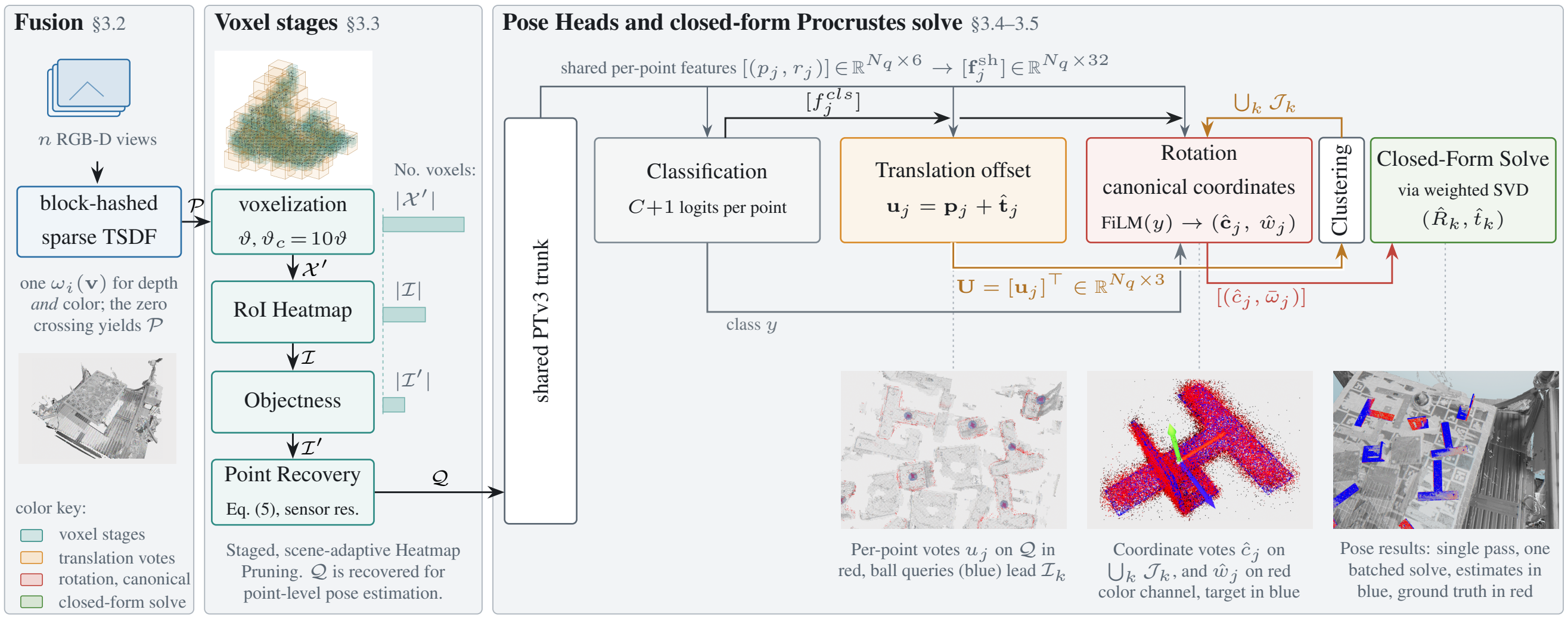
\begin{figure*}[t]
  \centering
  \input{sec/tikz_figures/fig_architecture1}
  \caption{\textbf{SCCV-6D.} $n$ RGB-D views are fused into a sparse TSDF, yielding $\mathcal P$. Two sparse, voxel-based heatmap stages focus task-relevant foreground regions $\mathcal I'$, before recovering to point-level resolution $\mathcal Q$. A shared PTv3 trunk feeds all three estimation heads. The translation votes $u_j$ naturally form clusters. Along class predictions, these gate a FiLM-conditioned canonical-coordinate head, and one single batched symmetry-aware weighted Procrustes solve returns 6D poses $[(\hat{R}_k, \hat{t}_k)]$.}
  \label{fig:arch}
\end{figure*}

\subsection{Problem Setting and Notation}
\label{sec:overview}

Given $n$ synchronized color and depth observations $\{(\mathcal{C}_i,\mathcal{D}_i)\}_{i=1}^{n}$ with known intrinsics $\mathcal{K}_i$ and extrinsics $\mathcal{T}_i$, we estimate the rigid transformations $(\bR,\bt)\in\SE$ that map the local frame of every object of a set of known classes into the world frame. The number of target objects is unknown and may be arbitrary. All object poses of a scene are produced in a single forward pass over the fused observation, which is what we refer to as holistic processing. No pose is estimated per crop, and no stage is executed once per candidate. \Cref{fig:arch} gives an overview of SCCV-6D.

\noindent The canonical model point cloud $\mathcal{M}_y\subset\R^3$ of every class $y\in\{1,\dots,C\}$ is available and centered at its origin. We assume the per-class symmetry group $\mathcal{G}_{k}$ to be known. For a ground-truth object $k$ of class $y_k$ with pose $(\bR^\ast_k,\bt^\ast_k)$ let
\begin{equation}
  \mathcal{O}_k=\{\bR^\ast_k\bm+\bt^\ast_k \mid \bm\in\mathcal{M}_{y_k}\}
  \label{eq:targetobj}
\end{equation}
be the target object, that is the model cloud placed in the world frame by its ground-truth pose. All training targets derive from~\Cref{eq:targetobj}. Throughout, 
\begin{equation}
    \kappa_\vartheta(\bp)=\lfloor(\bp-\bp_{\min})/\vartheta\rfloor\in\mathbb{N}^3
\end{equation}

\noindent denotes the index of a voxel with edge length $\vartheta$ containing the point $\bp$. The logistic function is denoted as $\sigma$.

\subsection{Multi-View RGB-D Fusion}
\label{sec:fusion}

The views are integrated into a sparse truncated signed distance field~\cite{curless1996volumetric,newcombe2011kinectfusion} held in a block-hashed grid~\cite{zhou2018open3d}. For a voxel center $\bv$ inside an active block, view $i$ contributes the truncated signed distance
\begin{equation}
  \phi_i(\bv)=\mathrm{clamp}\!\left(s_i(\bv)/\tau_{\mathrm{sdf}},-1,1\right), 
  s_i(\bv)=\mathcal{D}_i(\rho_i)-z ,
\end{equation}
with $\rho_i=\pi(\mathcal{K}_i\mathcal{T}_i^{-1}\bv)$ the pixel onto which $\bv$ projects in view $i$ and $(x,y,z)^\top=\mathcal{T}_i^{-1}\bv$ its coordinates in that camera frame. The binary visibility weight $\omega_i(\bv)$ is one if $\bv$ projects inside the image, carries a valid depth measurement and lies within the truncation band, and zero otherwise. Geometry $\Phi$ and appearance $\Psi$ are accumulated as
\begin{equation}
  \Phi(\bv)=\frac{\sum_i\omega_i(\bv)\phi_i(\bv)}{W(\bv)},\quad
  \Psi(\bv)=\frac{\sum_i\omega_i(\bv)\,\mathcal{C}_i(\rho_i)}{W(\bv)},
  \label{eq:colour}
\end{equation}
with $W(\bv)=\sum_i\omega_i(\bv)$. Sharing the weights keeps the two fields consistent, so a view that contributes an unreliable depth measurement also contributes little color.

\noindent Extracting the zero crossings of $\Phi$ and reading $\Psi$ at those positions yields the colored 3D point cloud $\mathcal{P}=\{(\bp_j,\br_j)\}_{j=1}^{N}$ with $\bp_j\in\R^3,\ \br_j\in[0,1]^3$. Averaging over views suppresses much of the speckle and view-dependent dropout on specular parts \cite{henry_iros, yang2021robi}.

\subsection{Staged Heatmaps and Point Restoration}
\label{sec:heatmaps}

The two heatmap stages follow~\cite{leuze2026sdt6d} and are summarized to set notation. We voxelize $\mathcal{P}$ with voxel size $\vartheta$, which yields the attributed sparse set $\mathcal{X}=\{(\bv_i,\bff_i)\}$ and build a coarse clone $\mathcal{X}'$ with $\vartheta_c=10\;\vartheta$. A submanifold sparse U-Net~\cite{graham2018submanifold} predicts region-of-interest scores $\hat h_i\in[0,1]$ on $\mathcal{X}'$ under a Gaussian focal loss~\cite{law2018cornernet}, with the target heatmap peaking at object centroids and boundaries. The soft attention prior $a_i=\sigma(\beta_a(\hat h_i-\epsilon_a))$ is thresholded at $\kappa_h$, and every fine voxel whose coarse clone is retained is kept. This gives the high-resolution set $\mathcal{I}$ in a scene-adaptive manner. A second sparse module predicts objectness $\hat h_{\mathrm{obj},i}$ on $\mathcal{I}$ under a focal loss~\cite{lin2017focal}. We prune by score rank, retaining the highest‑scoring half to form the object‑centric set $\mathcal{I}'$.

\noindent A voxel of edge length $\vartheta$ holds one feature for all points it contains, so prediction made on $\mathcal{I}'$ ~\cite{leuze2026sdt6d} carries an encoding error of order $\vartheta$. Let $\mathcal{V}'=\{\kappa_\vartheta(\bp)\mid(\bp,\cdot)\in\mathcal{I}'\}$ be the index set of the voxels. We restore to point-level with
\begin{equation}
  \mathcal{Q}=\{(\bp_j,\br_j)\in\mathcal{P}\mid \kappa_\vartheta(\bp_j)\in\mathcal{V}'\} ,
  \quad |\mathcal{Q}|=N_q \;,
  \label{eq:recovery}
\end{equation}
with $N_q$ being dynamic, given the scene-adaptive properties. This containment check costs one sort and no neighborhood query. Its inverse map is stored in-line and reused for targeting and for transporting attention priors in Sec.~\ref{sec:kd}. 

\subsection{Shared Trunk, Per-Point Classification and Translation-Offset Voting}
\label{sec:heads}

Every restored point of $\mathcal{Q}$ is described by $(\bp_j,\br_j)\in\R^6$. A shared PTv3 encoder~\cite{wu2024ptv3} maps $\mathcal{Q}$ to features $\bff^{\mathrm{sh}}_j\in\R^{32}$. All three heads need the same description of the local surface and its surrounding clutter, so a shared trunk avoids triplicating the cost. A second encoder refines these features for classification $\bff^{\mathrm{cls}}_j \in \R^{32}$, and an MLP predicts per-point logits over $C+1$ classes. Point $\bp_j$ is labeled with class $y_k$ if it lies inside the voxelized occupancy (cf. Eq.~\ref{eq:recovery}) of the target object $\mathcal{O}_k$, i.e. if $\kappa_{\vartheta}(\bp_j)\in\kappa_{\vartheta}(\mathcal{O}_k)$, and with background class $0$ otherwise, which is required because the objectness stage prunes conservatively and retains points on no target object.

\noindent Translation and rotation heads both start from $[(\bp_j,\br_j),\bff_j^{sh},\bff_j^{cls}]$, projected linearly to $32$ channels. Each head applies two PTv3 \cite{wu2024ptv3} encoders of different depth in sequence. PTv3 attends over patches of fixed token count taken from a serialized ordering of the points, so the receptive field is controllable by grid density. Each pooling step doubles the spatial extent at constant attention cost. The first encoder pools twice and keeps patches close to the surface, encoding the fine geometric evidence. The second pools four times and reaches local context. The token count is decimated accordingly, so the context is obtained at no additional memory. Where~\cite{leuze2026sdt6d} obtains fidelity and reach from a dual-branch transformer block of differing window sizes, we arrange the two sequentially, so the coarse encoder operates on features that already encode fine geometry, leaving the recombination of the two scales to the decoder skip connections. 

\noindent For every point $\bp_j$, the translation head predicts its relative offset $\hat\bt_j$ to the object centroid under a smooth $\ell_1$ loss. The vote of a point is $\bu_j=\bp_j+\hat\bt_j\in\R^3$. The target $t_j$ follows the voxelized occupancy regime already used for classification, with $t_j=0$ for $\kappa_{\vartheta}(\bp_j)\notin\kappa_{\vartheta}(\mathcal{O}_k)$. The per-point voting formulation sharpens the density peaks that the clustering in Sec.~\ref{sec:canonical} relies on.

\subsection{Per-Point Canonical Coordinate Voting}
\label{sec:canonical}

\paragraph{Object formation.} Votes $\bU=[\bu_1,\dots,\bu_{N_q}]^\top\in\R^{N_q\times3}$ concentrate near object centroids by construction, forming clusters without any further mechanism. A single DBSCAN~\cite{ester1996dbscan} pass returns $K$ cluster centers $\bq_1,\dots,\bq_K$, where $K$ is determined by the scene and not by any fixed budget. A point is recovered as a member of cluster $k$ if its own vote falls inside a ball of radius $\rho_c$\, around $\bq_k$,
\begin{equation}
  \ell_j=
  \begin{cases}
    \arg\min_k\|\bu_j-\bq_k\|_2, & \min_k\|\bu_j-\bq_k\|_2\le\rho_c\\[1pt]
    -1, & \text{otherwise}
  \end{cases}
  \label{eq:assign}
\end{equation}
with $\boldsymbol{\ell}\in\{-1,0,\dots,K-1\}^{N_q}$ and $\mathcal{J}_k=\{j\mid\ell_j=k\}$. The class of an object is the mode of the point-wise class predictions inside $\mathcal{J}_k$. Points with $\ell_j=-1$ do not enter the rotation head. The sequence of voting, clustering and membership recovery provides a task-driven sparsification scheme that yields sharper point cloud segments. This is a substantive difference to~\cite{leuze2026sdt6d}, where the pose is predicted on the retained grid segments and inherits the uncertainty of the heatmap stages. We find the additional sharpening to benefit pose accuracy directly.

\paragraph{Canonical Coordinate Voting.} Let $j\in\mathcal{J}_k$ and let $y$ be the class of object $k$. The rotation head predicts $\hat\bc_j\in\R^3$ and  $\hat w_j\in[0,1]$. With $\hat\bc_j$ the coordinate of $\bp_j$ in the canonical frame of $\mathcal{M}_y$. The pair $(\bp_j,\hat\bc_j)$ is therefore a 3D-3D correspondence between the reconstruction and the model. The confidence score $\hat w_j$ states how reliable the head considers that correspondence to be. A rigid transform that maps $\hat\bc_j$ to $\bp_j$ for all $j$ is the pose of object $k$. 

\paragraph{Head.} Rotation features are encoded by the cascade of Sec.~\ref{sec:heads} restricted to $\bigcup_k\mathcal{J}_k$, which reuses the shared and classification features already available. With $\bff_j\in\R^{64}$, an MLP trunk $\varphi_\theta$ is modulated by feature-wise linear modulation~\cite{perez2018film} conditioned on the predicted class $y$. Two linear maps produce coordinates and confidences,
\begin{align}
  \mathbf{h}_j &= \varphi_\theta(\bff_j), &
  \tilde{\mathbf{h}}_j &= \gamma(y)\odot\mathbf{h}_j+\beta(y), \label{eq:film}\\
  \hat\bc_j &= \bW_c\tilde{\mathbf{h}}_j+\mathbf{b}_c, &
  \hat w_j &= \sigma(\mathbf{w}_w^{\!\top}\tilde{\mathbf{h}}_j+b_w). \label{eq:head}
\end{align}
Modulation instead of per-class heads keeps the parameter count independent of $C$.

\paragraph{Training Targets.} For object $k$, the observed point is mapped into the canonical frame by the ground-truth pose and matched to the model,
\begin{equation}
  \tilde\bp_j=\bR^{\ast\top}_k(\bp_j-\bt^\ast_k),\quad
  \bc_j^\ast=\arg \hspace{-0,225cm} \min_{\bm\in\mathcal{M}_{y}}\|\tilde\bp_j-\bm\|_2 ,
  \label{eq:target}
\end{equation}
with residual $d_j=\|\tilde\bp_j-\bc^\ast_j\|_2$. We define a valid correspondence if $d_j\le\tau_{d}$\, which removes reconstruction outliers and points misassigned by the clustering. The confidence target is the soft indicator $\bar\omega_j=\sigma(-2\;(d_j-\tau_{d}/2))$, with $d_j$ in millimeters, so confidence expresses model‑surface agreement instead of a binary inlier classification.

\paragraph{Objective.} Let the discrete symmetries for object $k$ be defined by a finite group $\mathcal{G}_{k}\subset\SO$, where $\mathcal{G}_{k}=[\mathbf{I}]$ for asymmetric parts. With stop-gradient normalized weights $\bar w_j=\mathrm{sg}[\hat w_j]/\sum_{i\in\mathcal{J}_{k}}\mathrm{sg}[\hat w_i]$ and $\mathcal{H}_\delta$ the Huber loss over coordinates, the coordinate loss of object $k$ is
\begin{equation}
  \mathcal{L}^{k}_{\mathrm{coord}}=\min_{\bS\in\mathcal{G}_{k}}\ \sum_{j\in\mathcal{J}_{k}}\bar w_j\,
  \mathcal{H}_\delta\!\left(\hat\bc_j,\ \bS\,\bc^\ast_j\right).
  \label{eq:coord}
\end{equation}
We detach the confidence weights, to prevent the weighted mean from being minimized by redistributing weights rather than reducing residuals. Confidences are supervised separately by a binary cross-entropy $\mathcal{L}_{\mathrm{conf}}$ against $\bar\omega_j$. 

\noindent The correspondence formulation allows a trivial collapse to the canonical origin, especially for small objects. We add a spread term penalizing insufficient per-axis variance, $\mathcal{L}^{k}_{\mathrm{spread}}=\tfrac{1}{3}\sum_{d}\max(0,\varsigma_{\min}^{(y)}-\mathrm{Var}_{j\in\mathcal{J}_{k}}[\hat c_{j,d}])$, with $\varsigma_{\min}^{(y)}$ derived offline from the model's smallest per-axis variance, and a pairwise distance term $\mathcal{L}^{k}_{\mathrm{pair}}$ that compares predicted canonical distances to observed ones. For $n_p$ pairs $(u,v)$ sampled inside the cluster, let the objective be defined as $\mathcal{L}^{k}_{\mathrm{pair}}=1/{n_p}\textstyle\sum_{uv}\mathcal{H}_{\delta}\!\left(d^{\mathrm{can}}_{uv},d^{\mathrm{obs}}_{uv}\right)$, with $d^{\mathrm{can}}_{uv}=\|\hat\bc_u-\hat\bc_v\|_2$, and $d^{\mathrm{obs}}_{uv}=\|\bp_u-\bp_v\|_2$. The pose objective is defined as
\begin{equation}  \mathcal{L}_{\mathrm{pose}}=\lambda_c\mathcal{L}_{\mathrm{coord}}+\lambda_w\mathcal{L}_{\mathrm{conf}}
  +\lambda_s\mathcal{L}_{\mathrm{spread}}+\lambda_p\mathcal{L}_{\mathrm{pair}} \;.
  \label{eq:pose}
\end{equation}
\paragraph{Closed-form recovery.} At inference, the votes are gated by threshold $\tau_w$, as $w_j=\hat w_j \mathbf{1}[\hat w_j\ge\tau_w]$. Let $\bar\bp$ and $\bar\bc$ denote the weighted means of the observed points $p_j$ and of their predicted canonical coordinates $\hat{c}_j$.The pose minimizing $\sum_j w_j\|\bp_j-(\bR\hat\bc_j+\bt)\|_2^2$ is obtained by the SVD of the cross-covariance $\bH=\sum_{j}\bar w_j(\hat\bc_j-\bar\bc)(\bp_j-\bar\bp)^{\!\top}=\bU\boldsymbol{\Sigma}\bV^{\!\top}$ as the orthogonal Procrustes solution~\cite{kabsch1976,horn1987closed}
\begin{equation}
  \hat\bR_{k}=\bV\,\mathrm{diag}\!\left(1,1,\det(\bV\bU^{\!\top})\right)\bU^{\!\top},
  \hspace{0.115cm} \hat\bt_{k}=\bar\bp-\hat\bR_{k}\bar\bc .
  \label{eq:procrustes}
\end{equation}
The solve holds gradients in $\{\hat\bc_j,w_j\}$, so a direct pose loss could be added without architectural change. Our implementation costs one $3\times3$ SVD per object $k$, batched over the whole scene. Translation is re-estimated jointly with rotation and is therefore consistent with it. 

\subsection{Heatmap-Guided Feature Distillation}
\label{sec:kd}

The heatmap formulation produces attention priors that are supervised for the pose task, which we exploit in an on-top distillation module. Teacher $T$ and student $S$ share the grid construction, the staged network design, and differ only in capacity. The student uses shallower sparse U-Nets in the heatmap stages, and fewer transformer blocks with reduced channel dimensions in the encoders.

\noindent Both models prune adaptively, so their active sites do not coincide, and a feature loss cannot be indexed by position. Let $\mathcal{A}^T_s$ and $\mathcal{A}^S_s$ be the active sites of teacher and student at stage $s$. Further, let $\eta_s(\cdot)$ be the identity of a voxel at voxel-stages or point, which is the index $\kappa_{\vartheta_s}(\cdot)$ through Eq.~\eqref{eq:recovery}. The matching
\begin{equation}
  m_s(i)=
  \begin{cases}
    i', & \exists\,i'\in\mathcal{A}^T_s:\ \eta_s(i')=\eta_s(i)\\[1pt]
    \varnothing, & \text{otherwise}
  \end{cases}
  \label{eq:match}
\end{equation}
is unique, because $\eta_s$ is injective on the active sites of either model, and is obtained by one hash lookup per student site. The transfer weight is the attention prior $\pi^T$ the teacher assigns,
\begin{equation}
  \alpha^s_i=
  \begin{cases}
    \pi^{T}_{s,m_s(i)}, & m_s(i)\neq\varnothing\\
    0, & m_s(i)=\varnothing
  \end{cases}
  \label{eq:alpha}
\end{equation}
with $\pi^T_s$ equal to $a$ at the region-of-interest stage, to $\hat h_{\mathrm{obj}}$ at the objectness stage (Sec. \ref{sec:heatmaps}), and $\hat h'_{\mathrm{obj}}$ lifted through Eq.~\eqref{eq:recovery} at the point stages. We hypothesize that the teacher separates object-
centric segments more sharply than the student. A student site the teacher discarded is treated as background. With a linear projection $\bP_s$ that aligns channel widths, let the objective at stage $s$ be
\begin{equation}
  \mathcal{L}^{s}_{\mathrm{KD}}=\frac{\sum_{i\in\mathcal{A}^S_s}\alpha^s_i\big\|\bff^{T}_{m_s(i)}-\bP_s\bff^{S}_i\big\|_2^2}{\sum_{i\in\mathcal{A}^S_s}\alpha^s_i+\varepsilon_{kd}},
  \label{eq:kd}
\end{equation}
applied at the region-of-interest backbone, the objectness backbone, the shared trunk, and the second PTv3 encoder blocks for translation and rotation. Compared with an occupancy-weighting scheme ~\cite{zhang2023pointdistiller}, the weights of Eq.~\eqref{eq:alpha} are task-supervised. The complete objective combines the staged losses of the localization path with Eq.~\eqref{eq:pose},
\begin{equation}
  \mathcal{L}=\lambda_1\mathcal{L}_{\mathrm{RoI}}+\lambda_2\mathcal{L}_{\mathrm{obj}}+\lambda_3\mathcal{L}_{\mathrm{cls}}
  +\lambda_4\mathcal{L}_{\bt}+\mathcal{L}_{\mathrm{pose}} ,
\end{equation}
extended by $\sum_s\mu_s\mathcal{L}^{s}_{\mathrm{KD}}$ when a student is trained. Memory and compute follow the active sites of each stage, and the only quantities that scale with the number of objects are the DBSCAN pass and the batched decomposition.

%% file: sec/tikz_figures/fig_architecture1.tex
\definecolor{sccvInk}    {HTML}{1A1D21}
\definecolor{sccvSlate}  {HTML}{5A6672}
\definecolor{sccvMist}   {HTML}{AEB6BE}
\definecolor{sccvShell}  {HTML}{F2F4F6}
\definecolor{sccvBlue}   {HTML}{2C6FA6}
\definecolor{sccvTeal}   {HTML}{2E8B84}
\definecolor{sccvAmber}  {HTML}{D98C2B}
\definecolor{sccvCrimson}{HTML}{C0443C}
\definecolor{sccvGreen}  {HTML}{4B8B3B}
\definecolor{sccvGrey}   {HTML}{7E8A94}

\providecommand{\sccvttl}[1]{{\fontsize{7.6}{8.6}\selectfont\bfseries #1}}
\providecommand{\sccvbox}[1]{{\fontsize{7}{8}\selectfont #1}}
\providecommand{\sccvmic}[1]{{\fontsize{5.6}{6.4}\selectfont #1}}
\providecommand{\sccvsec}[1]{{\fontsize{6.2}{7}\selectfont\color{sccvSlate}\S#1}}

\ifdefined\mathbb
  \providecommand{\sccvR}{\mathbb{R}}
\else
  \providecommand{\sccvR}{\mathbf{R}}
\fi

\usetikzlibrary{backgrounds,arrows.meta}

\tikzset{
  sccvpanel/.style    = {rounded corners=1.6pt, fill=sccvShell, draw=sccvMist,
                         line width=0.35pt},
  sccvpanelttl/.style = {anchor=north west, inner sep=0pt, text=sccvInk},
  sccvblk/.style      = {rounded corners=1.4pt, draw=sccvSlate, line width=0.5pt,
                         fill=white, align=center, inner sep=2pt, text=sccvInk},
  sccvblkb/.style     = {sccvblk, draw=sccvBlue,    fill=sccvBlue!7},
  sccvblkt/.style     = {sccvblk, draw=sccvTeal,    fill=sccvTeal!8},
  sccvblka/.style     = {sccvblk, draw=sccvAmber,   fill=sccvAmber!10},
  sccvblkc/.style     = {sccvblk, draw=sccvCrimson, fill=sccvCrimson!7},
  sccvblkg/.style     = {sccvblk, draw=sccvGreen,   fill=sccvGreen!8},
  sccvflow/.style     = {draw=sccvInk, line width=0.55pt,
                         -{Stealth[length=1.9mm,width=1.4mm]}},
  sccvthin/.style     = {draw=sccvSlate, line width=0.4pt,
                         -{Stealth[length=1.5mm,width=1.1mm]}},
  sccvimg/.style      = {anchor=north west, inner sep=0pt, draw=sccvMist,
                         line width=0.45pt, dash pattern=on 1.6pt off 1.4pt,
                         fill=white, minimum width=2.5cm, minimum height=1.75cm,
                         align=center, text=sccvMist!70!black},
    sccvimg2/.style      = {anchor=north west, inner sep=0pt, draw=sccvMist,
                         line width=0.45pt, dash pattern=on 1.6pt off 1.4pt,
                         fill=white, minimum width=1.75cm, minimum height=1.275cm,
                         align=center, text=sccvMist!70!black},
  sccvnote/.style     = {font=\fontsize{5.7}{6.7}\selectfont, text=sccvSlate,
                         align=flush center, anchor=north, inner sep=1pt},
}

\begin{tikzpicture}[x=1cm,y=1cm]

\def\PB{0.60}\def\PT{7.35}                  %
\def\PIa{0.00}\def\PIb{2.10}                %
\def\PIIa{2.22}\def\PIIb{5.30}              %
\def\PIIIa{5.42}\def\PIIIb{17.30}           %
\def\SPINE{4.95}                            %

\def\colA{7.80}\def\colB{10.53}\def\colC{13.26}\def\colD{15.99}
\def\imgA{6.55}\def\imgB{9.28}\def\imgC{12.01}\def\imgD{14.74}
\def\imgY{3.30}                             %
\def\blkY{5.30}                             %
\def\blkT{5.88}\def\blkB{4.72}              %

\begin{scope}[on background layer]
  \draw[sccvpanel] (\PIa,\PB)   rectangle (\PIb,\PT);
  \draw[sccvpanel] (\PIIa,\PB)  rectangle (\PIIb,\PT);
  \draw[sccvpanel] (\PIIIa,\PB) rectangle (\PIIIb,\PT);
\end{scope}

\node[sccvpanelttl] at (0.10,7.27)  {\sccvttl{Fusion}~\sccvsec{3.2}};
\node[sccvpanelttl] at (2.32,7.27)  {\sccvttl{Voxel stages}~\sccvsec{3.3}};
\node[sccvpanelttl] at (5.52,7.27)  {\sccvttl{Pose Heads and closed-form Procrustes solve}~\sccvsec{3.4--3.5}};

\foreach \i/\o in {0/0.000,1/0.055,2/0.110}{
  \draw[rounded corners=1pt, draw=sccvBlue, line width=0.45pt,
        fill=sccvBlue!\the\numexpr 6+6*\i\relax]
        (0.48+\o,6.12+\o) rectangle (1.28+\o,6.64+\o);
}
\draw[draw=sccvBlue!70, line width=0.35pt] (0.72,6.30) -- (0.92,6.50) -- (1.10,6.34);
\node[anchor=north, text=sccvSlate] at (1.03,6.06) {\sccvmic{$n$ RGB-D views}};

\draw[sccvflow] (1.03,5.62) -- (1.03,5.37);
\node[sccvblkb, minimum width=1.82cm, minimum height=0.78cm] (tsdf) at (1.05,\SPINE)
  {\sccvbox{block-hashed}\\[-0.3ex]\sccvbox{sparse TSDF}};
\node[sccvnote, text width=1.94cm] at (1.05,4.41)
  {one $\omega_i(\mathbf v)$ for depth \emph{and} color; the zero crossing yields $\mathcal P$};

\node[anchor=north west, inner sep=0] at (0.15, 3.5)  {\includegraphics[width=1.75cm]{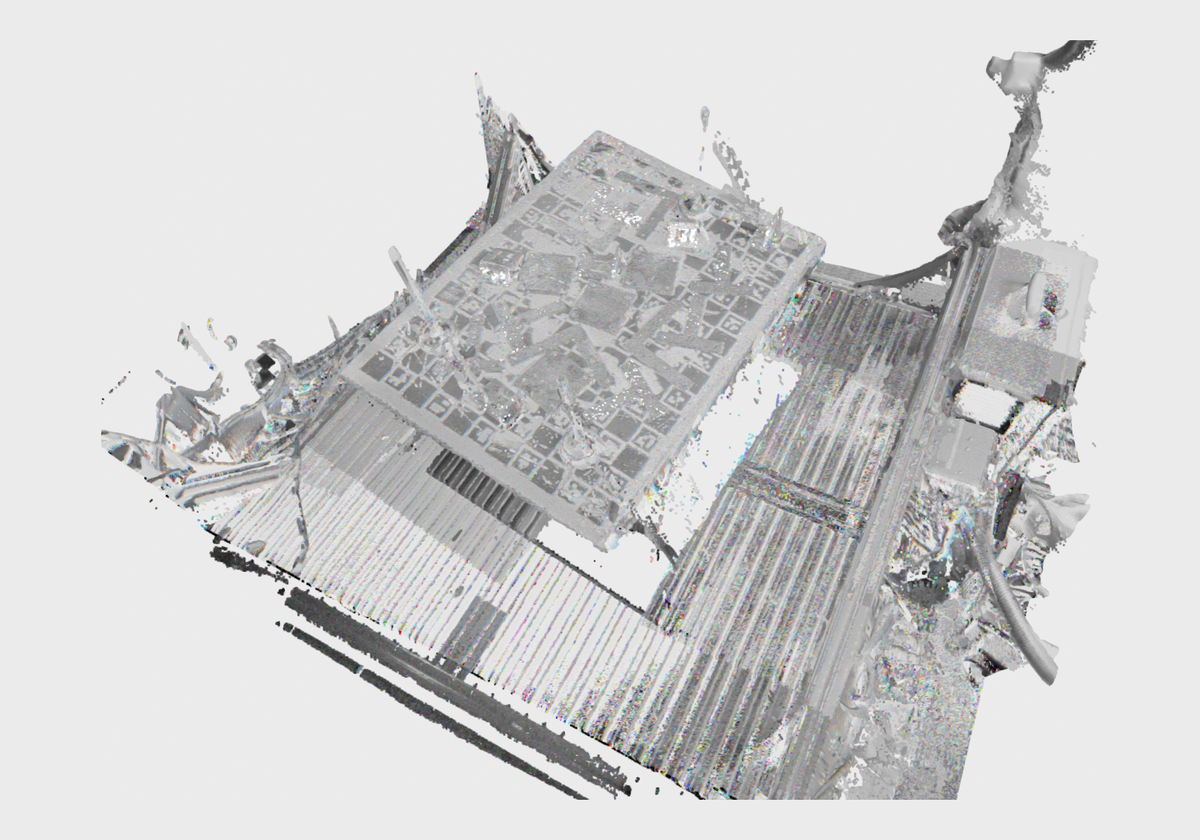}};

\def\KY{1.46}
\node[anchor=west, text=sccvSlate] at (0.0,\KY+0.30) {\sccvmic{color key:}};
\foreach \i/\c/\t in {0/sccvTeal/{voxel stages},
                      1/sccvAmber/{translation votes},
                      2/sccvCrimson/{rotation, canonical},
                      3/sccvGreen/{closed-form solve}}{
  \fill[\c!22, draw=\c, line width=0.3pt, rounded corners=0.5pt]
        (0.18,\KY-0.245*\i-0.055) rectangle (0.42,\KY-0.245*\i+0.075);
  \node[anchor=west, text=sccvSlate] at (0.48,\KY-0.245*\i+0.01) {\sccvmic{\t}};
}

\node[sccvblkt, minimum width=1.80cm, minimum height=0.68cm] (vox) at (3.20,\SPINE)
  {\sccvbox{voxelization}\\[-0.3ex]\sccvmic{$\vartheta$, $\vartheta_c\!=\!10\vartheta$}};
\node[sccvblkt, minimum width=1.80cm, minimum height=0.68cm] (roi) at (3.20,3.95)
  {\sccvbox{RoI Heatmap}};
\node[sccvblkt, minimum width=1.80cm, minimum height=0.68cm] (obj) at (3.20,2.95)
  {\sccvbox{Objectness}};
\node[sccvblkt, minimum width=1.80cm, minimum height=0.68cm] (res) at (3.20,1.95)
  {\sccvbox{Point Recovery}\\[-0.3ex]\sccvmic{Eq.~(\ref{eq:recovery}), sensor res.}};

\node[anchor=north west, inner sep=0] at (2.325, 6.85)  {\includegraphics[width=1.75cm]{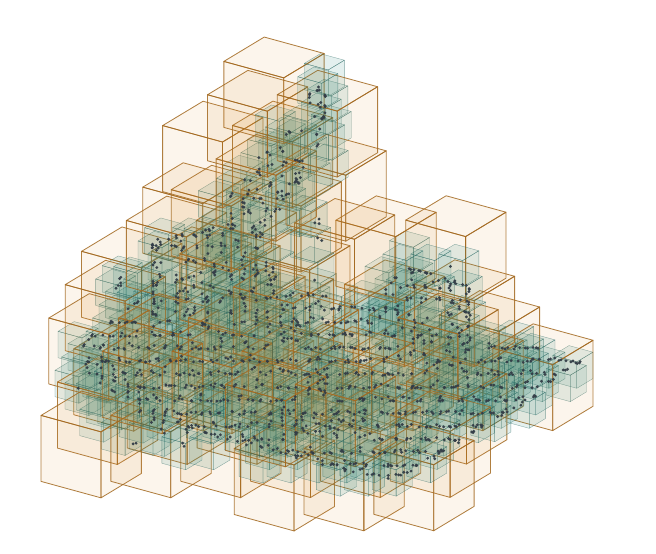}};

\draw[sccvflow] (tsdf.east) -- node[above=-0.5pt]{\sccvmic{$\mathcal P$}} (vox.west);
\draw[sccvflow] (vox.south) -- node[right=-1pt]{\sccvmic{$\mathcal X'$}} (roi.north);
\draw[sccvflow] (roi.south) -- node[right=-1pt]{\sccvmic{$\mathcal I$}}  (obj.north);
\draw[sccvflow] (obj.south) -- node[right=-1pt]{\sccvmic{$\mathcal I'$}} (res.north);

\node[anchor=south west, text=sccvSlate] at (4.20,5.34) {\sccvmic{No. voxels:}};
\draw[draw=sccvTeal!55, line width=0.35pt, dash pattern=on 1pt off 1pt]
      (4.20,5.10) -- (4.20,2.80);
\foreach \y/\w/\t in {4.95/0.90/{$|\mathcal X'|$}, 3.95/0.47/{$|\mathcal I|$},
                      2.95/0.24/{$|\mathcal I'|$}}{
  \draw[fill=sccvTeal!30, draw=sccvTeal!65, line width=0.3pt]
        (4.20,\y-0.11) rectangle (4.20+\w,\y+0.05);
  \node[anchor=south west, text=sccvSlate] at (4.20,\y+0.0) {\sccvmic{\t}};
}

\node[sccvnote, text width=2.85cm] at (3.76,1.42)
  {Staged, scene-adaptive Heatmap Pruning. $\mathcal{Q}$ is recovered for point-level pose estimation.};

\node[sccvblk, draw=sccvSlate, fill=white, minimum width=0.80cm,
      minimum height=4.50cm] (trunk) at (5.95,3.85) {};
\node[rotate=90, text=sccvInk] at (trunk.center) {\sccvbox{shared PTv3 trunk}};
\draw[sccvflow] (res.east) -- node[above=-1.5pt]{\sccvmic{$\mathcal Q$}} (5.55,1.95);

\draw[draw=sccvSlate, line width=0.5pt] (5.95,6.10) -- (5.95,6.45) -- (\colC-0.155,6.45);
\foreach \x in {\colA,\colB}{\draw[sccvthin] (\x,6.45) -- (\x,\blkT);}
\draw[sccvthin] (13.10,6.45) -- (13.10,\blkT);
\node[anchor=south west, text=sccvSlate] at (6.05,6.37)
  {\sccvmic{shared per-point features $\mathbf [(p_j, r_j)]\!\in\!\sccvR^{N_q\times6}\to[\mathbf f^{\mathrm{sh}}_j]\!\in\!\sccvR^{N_q\times32}$}};

\node[sccvblka, minimum width=2.50cm, minimum height=1.16cm] (htrn) at (\colB,\blkY)
  {\sccvbox{Translation offset}\\[-0.25ex]
   \sccvmic{$\mathbf u_j=\mathbf p_j+\hat{\mathbf t}_j$}};
\node[sccvblk, draw=sccvGrey, fill=sccvGrey!8, minimum width=2.50cm,
      minimum height=1.16cm] (hcls) at (\colA,\blkY)
  {\sccvbox{Classification}\\[-0.25ex]
   \sccvmic{$C{+}1$ logits per point}};
\node[sccvblkc, minimum width=2.50cm, minimum height=1.16cm] (hrot) at (\colC,\blkY)
  {\sccvbox{Rotation}\\[-0.3ex]\sccvbox{canonical coordinates}\\[-0.25ex]
   \sccvmic{FiLM$(y)\to(\hat{\mathbf c}_j,\ \hat w_j)$}};
\node[sccvblkg, minimum width=2.0cm, minimum height=1.16cm] (rec) at (\colD+0.2,\blkY)
  {\sccvbox{Closed-Form Solve}\\[-0.6ex]
   \sccvmic{via weighted SVD}\\[-0.6ex]\sccvmic{$(\hat{R}_k, \hat{t}_k)$}};
\node[sccvblkg, minimum width=1.16cm, minimum height=0.5cm, fill=white, rotate=90,draw=sccvSlate] (rec) at (\colD-1.15,\blkY){\sccvbox{Clustering}};

\draw[sccvflow, draw=sccvGrey!85!black]
      (\colA,\blkB) -- (\colA,3.95) -- (13.05,3.95) -- (13.05,\blkB);
\node[anchor=north, text=sccvGrey!85!black, inner sep=0.8pt] at (\colA+0.5,3.92)
  {\sccvmic{class $y$}};

\draw[sccvflow, draw=sccvGrey!25!black]
      (8.,\blkT) -- (8.,6.1) -- (10.5,6.1);
\node[anchor=north, text=sccvGrey!25!black, inner sep=0.8pt] at (9.20,6.47)
  {\sccvmic{$[f_j^{cls}]$}};
  \draw[sccvflow, draw=sccvGrey!25!black]
      (10.6,6.1) -- (13.08,6.1);

\draw[sccvflow, draw=sccvAmber!85!black,
      preaction={draw=white, line width=1.8pt}]
      (\colB,\blkB) -- (\colB,4.44) -- (\colD-1.15,4.44) -- (\colD-1.15,\blkB);
\node[anchor=south, text=sccvAmber!80!black, inner sep=0.8pt] at (\colB+1.2,4.05)
  {\sccvmic{$\bU=[\bu_j]^\top\in\R^{N_q\times3}$}};

\draw[sccvflow, draw=sccvAmber!85!black]
      (\colD-1.15,\blkT) -- (\colD-1.15,6.10) -- (13.35,6.10) -- (13.35,\blkT);
\node[anchor=south, text=sccvAmber!80!black, inner sep=0.8pt] at (14.0,6.12)
  {\sccvmic{$\bigcup_k\mathcal{J}_k$ }};

\draw[sccvflow, draw=sccvCrimson] (13.35,\blkB) -- (13.35,\blkB-0.45) -- (15.4, \blkB-0.45) -- (15.4, \blkB);
\node[anchor=south, text=sccvCrimson, inner sep=0.8pt] at (13.95,\blkB-0.78)
  {\sccvmic{$[(\hat{c}_j, \bar{\omega}_j)]$ }};

\begin{scope}[on background layer]
  \foreach \x in {\colB,\colC,\colD}{
    \draw[draw=sccvMist, line width=0.4pt, dash pattern=on 0.7pt off 1.0pt]
          (\x,\blkB) -- (\x,\imgY);}
\end{scope}

\node[anchor=north west, inner sep=0] at (\imgB,\imgY)  {\includegraphics[width=2.5cm]{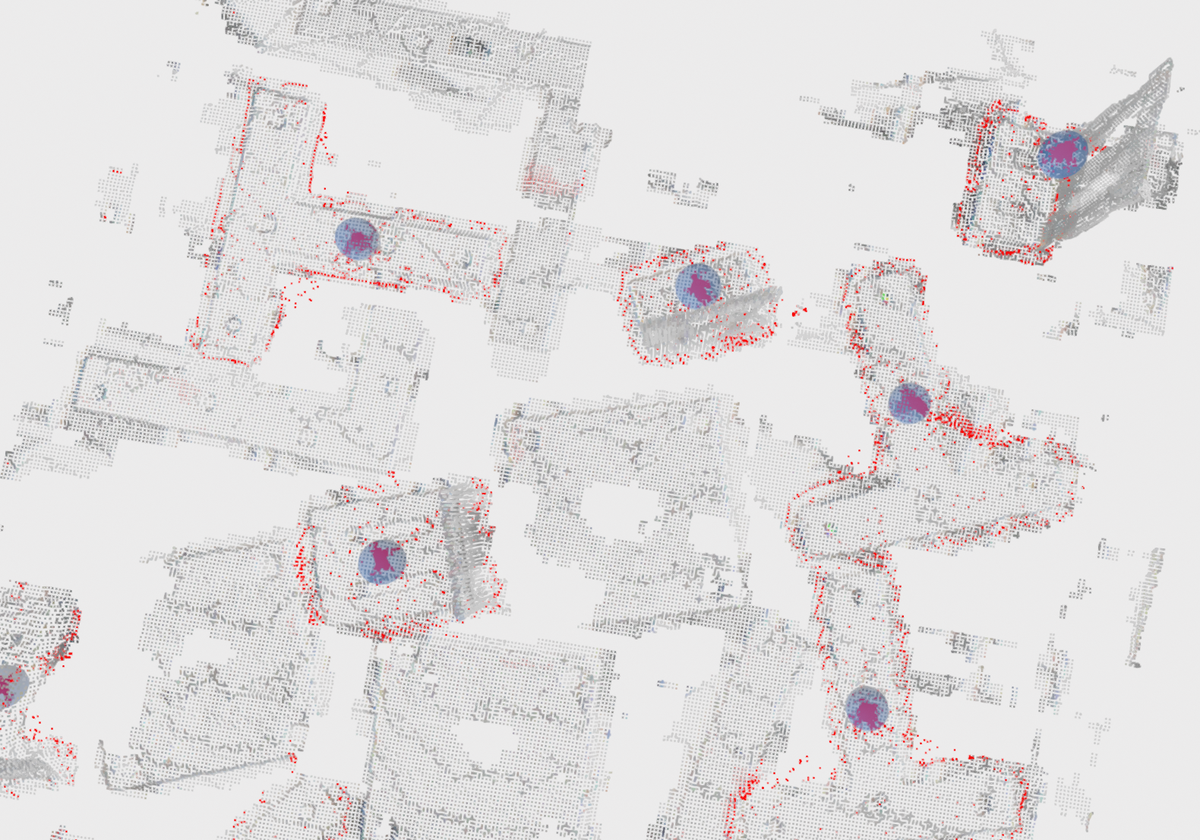}};
\node[anchor=north west, inner sep=0] at (\imgC,\imgY)  {\includegraphics[width=2.5cm]{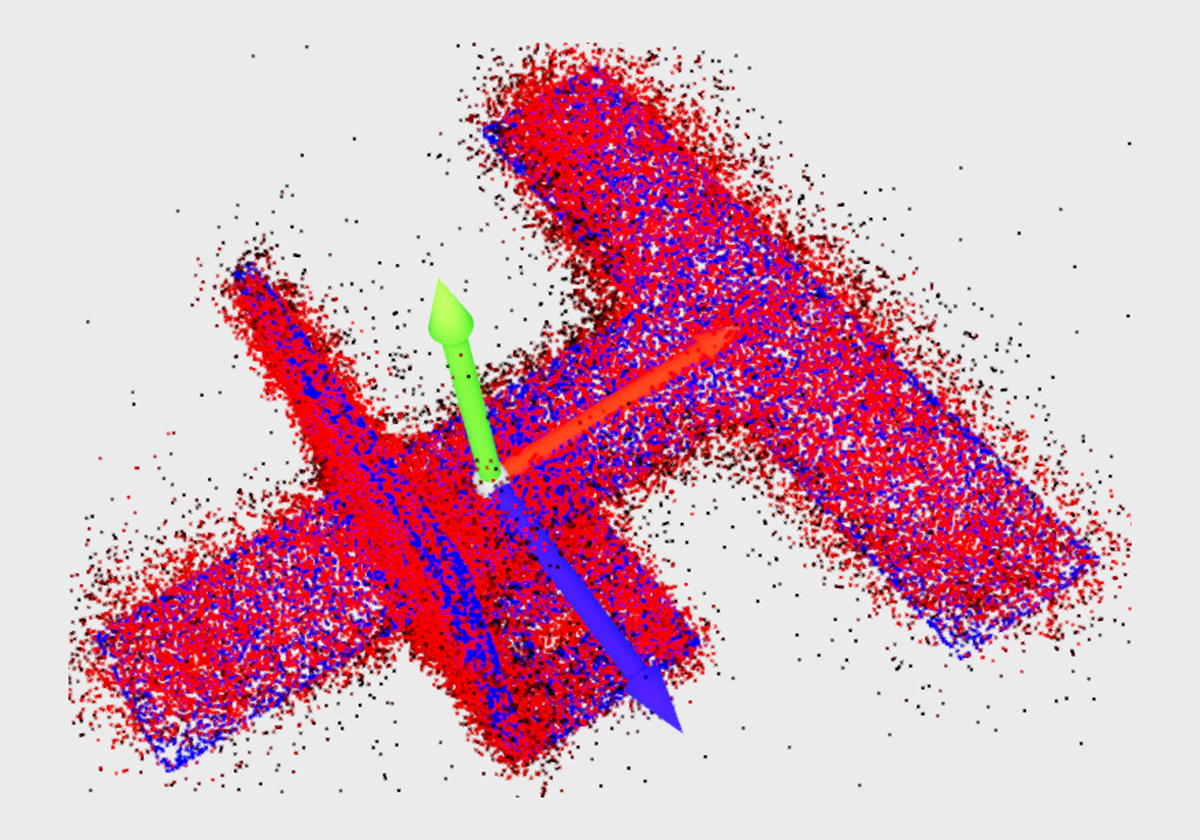}};
\node[anchor=north west, inner sep=0] at (\imgD,\imgY)  {\includegraphics[width=2.5cm]{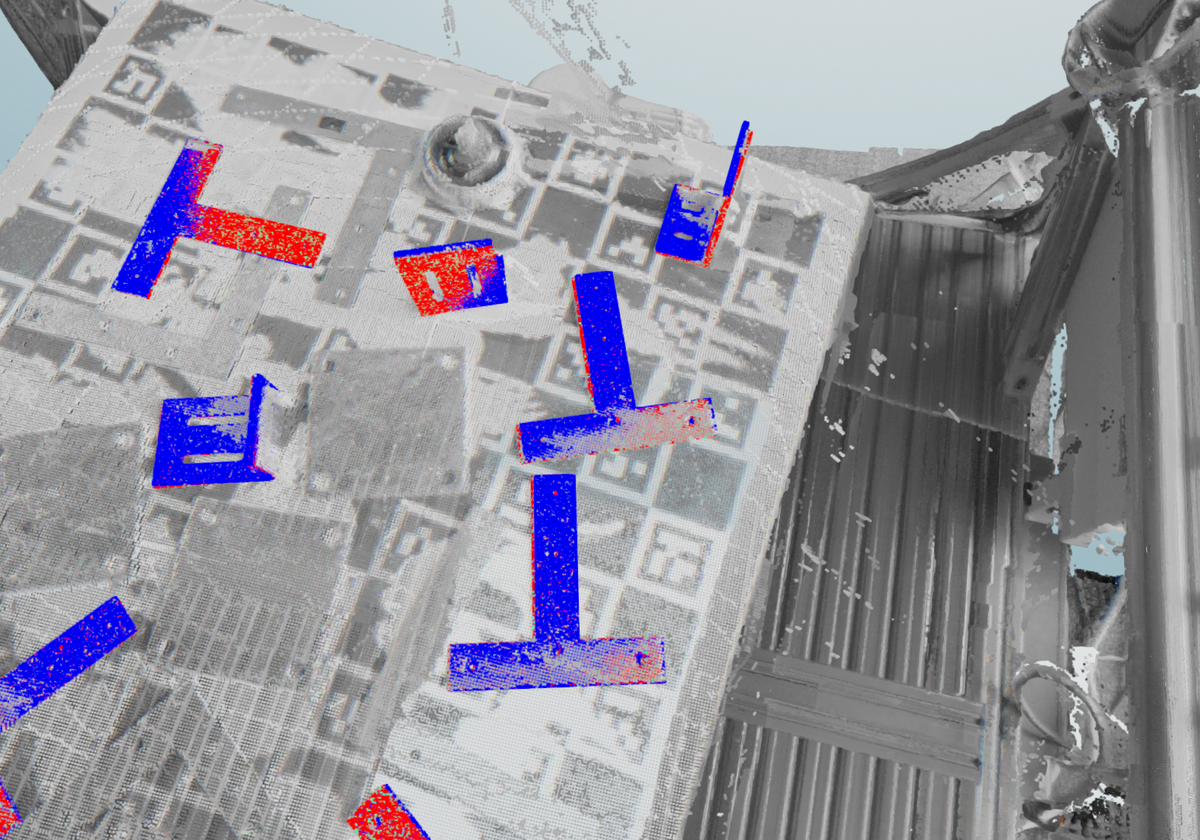}};

\node[sccvnote, text width=2.55cm] at (\colB,1.44)
  {Per-point votes $u_j$ on $\mathcal{Q}$ in red, ball queries (blue) lead $\mathcal{I}_k$};
\node[sccvnote, text width=2.55cm] at (\colC,1.44)
  {Coordinate votes $\hat{c}_j$ on $\bigcup_k\mathcal{J}_k$, and $\hat{w}_j$ on red \\ color channel, target in blue};
\node[sccvnote, text width=2.55cm] at (\colD,1.44)
  {Pose results: single pass, one batched solve, estimates in blue, ground truth in red};

\end{tikzpicture}

%% file: sec/4_results_experiments.tex
\section{Experiments}
\label{sec:experiments}

\subsection{Datasets, Metrics and Training}

We evaluate on three multi-view benchmarks covering complementary regimes. The Industrial Plenoptic Dataset (IPD)~\cite{kalra2024ipd} holds multi-instance scenes of ten industrial parts. We use four views and the official BOP protocol~\cite{nguyen2025bop}, reporting the average precision over the maximum symmetry-aware surface distance (MSSD) and projection distance (MSPD), the mean AP, and AP$^{\mathrm{mm}}_{\mathrm{MSSD}}$ at absolute millimeter thresholds. ROBI~\cite{yang2021robi} provides bins of seven reflective parts. We evaluate on the Ensenso test split with four views under the symmetry-aware ADD$^\ast$ against the numbers reported in~\cite{yang2025active}. MV-YCB-SymMovCam~\cite{duffhauss2023symfm6d} is a synthetic dataset of textured YCB objects~\cite{calli2015ycb}. We use three views for comparability with prior work \cite{duffhauss2022mv6d,duffhauss2023symfm6d, leuze2026sdt6d} and report the area under the ADD-S and ADD-(S) curves. 

\noindent \noindent All experiments run on a single NVIDIA RTX 5090. The heatmap heads are trained for two warm-up epochs, after which all heads are optimized jointly for $15$ epochs in total, with color and depth augmentations applied per view before fusion. During the first five epochs the cluster centers of Eq.~\eqref{eq:assign} are the ground-truth centroids, exposing the rotation head to well-recovered foreground while the translation head converges, after which the offset predictions replace them. The voxel size $\vartheta$ is $1.75$\,mm on IPD and MV-YCB and $0.5$\,mm on ROBI, whose capture volume is smaller, and the clustering scales with it, using $\varepsilon_{db}=8$\,mm and $\rho_c=10$\,mm on the former two and $3$\,mm and $5$\,mm on ROBI, at a minimum of $100$ samples. Loss weights are $(\lambda_1,\dots,\lambda_4)=(1,2,2,3)$ following~\cite{leuze2026sdt6d} and $(\lambda_c,\lambda_w,\lambda_s,\lambda_p)=(1.0,0.5,0.3,0.15)$. The remaining constants are set empirically, with $\beta_a=15$ and $\epsilon_a=0.5$ in the attention prior, $\tau_d=5$\,mm for a valid correspondence, $n_p=512$ sampled pairs, and $\tau_w=0.3$ at recovery. Averaged over the IPD split, reconstruction accounts for $0.837$\,s of the runtime, the forward pass including clustering and classification for $0.217$\,s, and the batched solve for $0.079$\,s, summing to the reported $1.13$\,s. Reconstruction is a GPU voxel-block-grid integration~\cite{zhou2018open3d} and follows the capture volume and the view count. The remaining $0.296$\,s cover the forward pass and the batched solve. 

\begin{table}[tb]
  \centering
  \footnotesize
  \setlength{\tabcolsep}{1.7pt}
  \begin{tabular}{lcccccc}
    \toprule
    & & AP & AP & AP$^{\mathrm{mm}}$ & AP & Time \\
    Method & AP & \scriptsize MSPD & \scriptsize MSSD & \scriptsize MSSD & \scriptsize 25 & [s] \\
    \midrule
    FRTPose-WAPRv2 (MV)~\cite{wang2025frtpose} & \textbf{.932} & \textbf{.968} & .897 & .661 & .897 & 33.9 \\
    3PT-Pose-Ind.~\cite{kalra2026threept}      & .923 & .933 & \textbf{.914} & \textbf{.822} & \textbf{.914} & 24.52 \\
    SEVD-0504~\cite{sevd2025}                  & .873 & .913 & .833 & .644 & .833 & 13.4 \\
    AlignPose~\cite{alignpose2025}             & .798 & .852 & .743 & .582 & .743 & -- \\
    MatchU~\cite{huang2024matchu}              & .520 & .566 & .475 & .340 & .475 & -- \\
    FRTPose-WAPRv2 (YOLO)~\cite{wang2025frtpose} & .746 & .761 & .731 & .593 & .731 & 1.34 \\
    FreeZeV2.2~\cite{caraffa2024freeze}        & .439 & .459 & .419 & .321 & .419 & 7.20 \\
    \midrule
    \methodname{} (ours)                       & .826 & .866 & .787 & .548 & .787 & \textbf{1.13} \\
    \bottomrule  
  \end{tabular}
  \caption{IPD test split under the BOP protocol~\cite{nguyen2025bop}. All competing entries are multi-stage systems relying on upstream detections and, in several cases, on ensembles of detectors and refiners. Competing runtimes are self-reported~\cite{nguyen2025bop} on heterogeneous hardware with differing modalities and view counts. }
  \label{tab:ipd}
\end{table}

\subsection{Results}

\paragraph{IPD.} \methodname{} reaches $0.826$ AP (Table~\ref{tab:ipd}), above AlignPose~\cite{alignpose2025} and the fastest competing configuration \cite{wang2025frtpose}, and below the leading multi-stage entries. 3PT-Pose-Industrial~\cite{kalra2026threept} is a pair of large transformers trained on web-scale object-centric data that refines its hypotheses over several passes in $24.52$\,s per scene. FRTPose \cite{wang2025frtpose} in the multi-view configuration couples a surface-coordinate network, a PnP solve, and a time-consuming refinement stage taking $33.9$\,s. \methodname{} produces the poses of all objects of a scene in $1.13$\,s from one forward pass and a batched decomposition, roughly $22\times$ and $30\times$ faster respectively, while retaining roughly $89\%$ of their AP. The only competing entry in the same latency regime, FRTPose with a YOLO front end, is $8.0$ points of AP below us at a comparable runtime. AlignPose~\cite{alignpose2025} reports no scene-level runtime on the leaderboard \cite{nguyen2025bop}, but the per-object refinement timings already exceed our full-scene latency.

\noindent The remaining gap is concentrated in AP$^{\mathrm{mm}}_{\mathrm{MSSD}}$, where 3PT leads by a wide margin. \cite{kalra2026threept} applies a render-and-refine stage that optimizes the pose directly against the observation and can polish an already correct hypothesis to sub-millimeter agreement. Our estimate is the closed-form optimum of a fixed set of learned correspondences and inherits the residual noise. We rank higher at the MSPD and MSSD thresholds, which reward finding and disambiguating an object over polishing it.

\begin{table}[tb]
  \centering
  \footnotesize
  \setlength{\tabcolsep}{2.1pt}
  \begin{tabular}{lccccc}
    \toprule
    & CosyPose & CosyPose & MV-3D-KP & MEC-Net & Ours \\
    & +PVNet & +LINE2D & \cite{li2023mvkp} & \cite{yang2025active} & \\
    Modality & RGB & RGB-D & RGB-D & RGB & RGB-D \\
    \midrule
    Tube Fitting    & 39.4 & 74.8 & \textbf{94.0} & 89.4 & 90.9 \\
    Chrome Screw    & 17.4 & 73.0 & \textbf{90.8} & 86.7 & 83.2 \\
    Eye Bolt        & 21.6 & 85.1 & \textbf{93.2} & \textbf{93.2} & 92.7 \\
    Gear            & 50.6 & 80.2 & 85.2 & \textbf{91.4} & 88.3 \\
    Zigzag          & 89.7 & 87.9 & \textbf{96.6} & 94.8 & 95.8 \\
    DIN Connector   & 13.3 & 57.8 & \textbf{90.6} & 69.5 & 79.5 \\
    D-Sub Connector & 11.2 & 55.3 & \textbf{92.5} & 79.5 & 89.5 \\
    \midrule
    Mean            & 34.7 & 73.4 & \textbf{91.8} & 86.4 & 88.6 \\
    \bottomrule
  \end{tabular}
  \caption{ROBI (Ensenso, four views) under ADD$^\ast$. Baselines as reported in~\cite{yang2025active}.}
  \label{tab:robi}
\end{table}

\begin{table}[tb]
  \centering
  \small
  \begin{tabular}{lcc}
    \toprule
    Method & ADD-S & ADD-(S) \\
    \midrule
    CosyPose~\cite{labbe2020cosypose}       & 92.3 & 87.7 \\
    MV6D~\cite{duffhauss2022mv6d}           & 91.2 & 85.6 \\
    SyMFM6D w/o sym.~\cite{duffhauss2023symfm6d} & 95.2 & 91.5 \\
    SyMFM6D~\cite{duffhauss2023symfm6d}     & 95.4 & 91.7 \\
    \prior~\cite{leuze2026sdt6d}            & 90.6 & 89.1 \\
    \midrule
    \methodname{} (ours)                    & \textbf{96.1} & \textbf{92.8} \\
    \bottomrule
  \end{tabular}
  \caption{MV-YCB-SymMovCam, three views, area under the ADD-S and ADD-(S) curves.}
  \label{tab:mvycb}
\end{table}

\paragraph{ROBI.} The second experiment probes the performance in densely packed bins of
small reflective parts (Table~\ref{tab:robi}). \methodname{} reaches $88.6$ mean
ADD$^\ast$, $2.2$ points above the RGB-only MEC-Net~\cite{yang2025active} and $3.2$ points
below MV-3D-KP~\cite{li2023mvkp}, which it leads on the Gear, the object carrying the
highest-order rotational symmetry of the set. The deficit concentrates on the chrome screw and the DIN connector, the two smallest parts. Rotation is observable only through the displacement it induces, so at a comparable correspondence error in millimeters the recoverable angle scales inversely with object size. MV-3D-KP crops and resizes every detection, normalizing resolution per object, and retains several pose hypotheses per detection that are individually refined against the depth map. Neither baseline reports scene-level runtimes. All of them are evaluated once per object and per view, whereas our poses follow from a single forward pass in 1.02\,s per scene. Latency is lower than on IPD despite the finer \(\vartheta\), attributable to the smaller capture volume that reduces the dominant sparse TSDF reconstruction cost.

\paragraph{MV-YCB-SymMovCam.} On the household benchmark (Table~\ref{tab:mvycb})
\methodname{} improves over \prior~\cite{leuze2026sdt6d} by $5.5$ points ADD-S and $3.7$
points ADD-(S) under an otherwise identical protocol. Fusing color into the
reconstruction exploits information that depth-only pipelines discard entirely on textured
objects. Our approach further outperforms the prior leading entry SyMFM6D~\cite{duffhauss2023symfm6d}, which fuses appearance and geometry throughout and uses symmetry-aware keypoint targets, by the narrower margin of $0.7$\,ADD-S and $1.1$\,ADD-(S). 

\begin{table}[tb]
  \centering
  \footnotesize
  \setlength{\tabcolsep}{4pt}
  \begin{tabular}{lcc}
    \toprule
    Configuration & AP & AP$^{\mathrm{mm}}_{\mathrm{MSSD}}$ \\
    \midrule
    Per-voxel rotation, batched ICP        & 0.712 & 0.468 \\
    w/o color (depth-only heads)         & 0.769 & 0.483 \\
    w/o point recovery (per-cell)          & 0.812 & 0.522 \\
    Single five-stage encoder per head     & 0.814 & 0.517 \\
    unweighted Procrustes solve            & 0.816 & 0.542 \\
    \midrule
    \methodname{} (full)                   & 0.826 & 0.548 \\
    \bottomrule
  \end{tabular}
  \caption{Component study on the IPD test split. Each row replaces or removes a single component of the full model. Effects are measured independently and are not additive.}
  \label{tab:ablation}
\end{table}

\begin{table}[tb]
  \centering
  \footnotesize
  \setlength{\tabcolsep}{3pt}
  \begin{tabular}{lccccc}
    \toprule
    & & AP & AP & AP$^{\mathrm{mm}}$ & Time \\
    Model & AP & \scriptsize MSPD & \scriptsize MSSD & \scriptsize MSSD & [s] \\
    \midrule
    Teacher                                                        & 0.841 & 0.884 & 0.798 & 0.552 & 1.11 \\
    \midrule
    $-25\%$, w/o weighting                                         & 0.723   & 0.781 & 0.665 & 0.485 & 1.02 \\
    $-25\%$, occupancy~\cite{zhang2023pointdistiller}              & 0.768 & 0.829 & 0.707 & 0.505 & 1.03 \\
    $-25\%$, heatmap (ours)                                        & \textbf{0.827} & \textbf{0.866} & \textbf{0.789} & \textbf{0.542} & 1.06 \\
    $-35\%$, occupancy~\cite{zhang2023pointdistiller}              & 0.747 & 0.821 & 0.673 & 0.504 & 1.03 \\
    $-35\%$, heatmap (ours)                                        & 0.810 & 0.849 & 0.771 & 0.523 & 1.01 \\
    \bottomrule
  \end{tabular}
  \caption{Distillation Results on the IPD validation set. Percentages denote the parameter reduction with respect to the teacher. The teacher is the full model of Table \ref{tab:ipd}.}
  \label{tab:kd}
\end{table}

\subsection{Distillation Study}
We probe the student performance on the IPD validation split. Teacher and students hold $126.7$ M, $95.1$ M and $82.3$ M parameters.  A student with $25\%$ fewer parameters trained under a uniformly averaged feature loss drops $11.8$ AP relative to the teacher (Table~\ref{tab:kd}). The loss arises in the heatmap stages, where foreground–background imbalance weakens relevance prediction. Downstream heads cannot recover foreground that was never retained. Voxel‑occupancy reweighting following PointDistiller~\cite{zhang2023pointdistiller} closes about one third of the gap. Our novel weighting strategy recovers $98\%$, leaving the student $1.4$ AP short and increasing the margin over \cite{zhang2023pointdistiller} from $5.9$ to $6.3$ AP at $35\%$ reduction. Our heatmaps remove most background before point‑level processing, so occupancy is computed on an object‑enriched but still noisy high‑resolution set. That noise is deliberate, carried by the Gaussian‑smoothed heatmap targets to retain contextual cues for cluttered objects, and it becomes overweighted under the occupancy‑based scheme and likely reduces the effectiveness.

\subsection{Component Study}
\label{sec:ablation}
Table~\ref{tab:ablation} isolates the individual design decisions. Replacing canonical
voting by the per-voxel Chamfer regression with batched ICP of~\cite{leuze2026sdt6d},
inside our otherwise unchanged pipeline, costs $11.4$ points AP and is by a wide margin the
largest single effect. Restricting the heads to depth reduces accuracy by $5.7$AP, the second largest effect. The loss reflects weaker separation rather than detection. The heatmap stages prune conservatively to retain contextual information. Appearance helps the classification and translation heads to distinguish retained background from clutter with similar geometry, producing sharper clusters and cleaner correspondences. Predicting on the voxel grid instead of the restored points reduces accuracy by $1.4$ points AP and by $2.6$ points AP$^{\mathrm{mm}}_{\mathrm{MSSD}}$. This pattern is consistent with an encoding error that distorts the pose estimate rather than the detection stage, and therefore impacts millimeter‑level metrics more strongly. The replacement of the two encoders of each head by a single one costs $1.2$ AP, and the removal of the confidence weighting a further $1.0$ AP. Confidence weighting contributes largely at the millimeter thresholds, where down-weighting unreliable correspondences matters most.

\paragraph{Limitations. } The method assumes a CAD model per class and a known, finite symmetry group, which suits industrial parts but excludes unseen objects. Continuous symmetries enter only through a discretization of that group, and a misspecified group is not detected at inference. Recovery further presumes extrinsics accurate relative to $\vartheta$, since an error of that order displaces the fused surface and the correspondences with it. Thresholds are set per dataset and scale with capture volume, as none of them is learned.

\section{Conclusion}

\noindent We presented \methodname, a fully sparse multi-view RGB-D 6D pose estimator that fuses color and depth into a sparse TSDF. Staged heatmaps focuses high-potential foreground regions in a scene-adaptive manner. The attention priors that drive the pruning are reused as transfer weights for feature distillation, which yields compact students that outperform density-based reweighting in this domain. SCCV-6D predicts class, translation and rotation for an arbitrary number of objects in a single forward pass. Rotation is cast as canonical coordinate voting, so every pose follows in closed form from a symmetry-aware weighted Procrustes solve. This removes the object-wise post-processing that otherwise scales with bin
occupancy. Across complementary regimes, we reported accuracy within reach of the leading multi-stage ensembles at significantly lower latency.

%% file: main.bbl
\begin{thebibliography}{57}
\providecommand{\natexlab}[1]{#1}
\providecommand{\url}[1]{\texttt{#1}}
\expandafter\ifx\csname urlstyle\endcsname\relax
  \providecommand{\doi}[1]{doi: #1}\else
  \providecommand{\doi}{doi: \begingroup \urlstyle{rm}\Url}\fi

\bibitem[{BOP Challenge Participants}(2025)]{sevd2025}
{BOP Challenge Participants}.
\newblock {SEVD-0504}.
\newblock \url{https://bop.felk.cvut.cz}, 2025.
\newblock BOP Challenge submission, accessed: 2026-08-18.

\bibitem[Brachmann et~al.(2014)Brachmann, Krull, Michel, Gumhold, Shotton, and
  Rother]{brachmann2014learning}
Eric Brachmann, Alexander Krull, Frank Michel, Stefan Gumhold, Jamie Shotton,
  and Carsten Rother.
\newblock Learning {6D} object pose estimation using {3D} object coordinates.
\newblock In \emph{European Conference on Computer Vision (ECCV)}, 2014.

\bibitem[Breyer et~al.(2021)Breyer, Chung, Ott, Siegwart, and
  Nieto]{breyer2021vgn}
Michel Breyer, Jen~Jen Chung, Lionel Ott, Roland Siegwart, and Juan Nieto.
\newblock Volumetric grasping network: Real-time {6 DOF} grasp detection in
  clutter.
\newblock In \emph{Conference on Robot Learning (CoRL)}, 2021.

\bibitem[Calli et~al.(2015)Calli, Singh, Walsman, Srinivasa, Abbeel, and
  Dollar]{calli2015ycb}
Berk Calli, Aaron Singh, Arjun Walsman, Siddhartha Srinivasa, Pieter Abbeel,
  and Aaron~M. Dollar.
\newblock The {YCB} object and model set: Towards common benchmarks for
  manipulation research.
\newblock In \emph{International Conference on Advanced Robotics (ICAR)}, 2015.

\bibitem[Caraffa et~al.(2024)Caraffa, Boscaini, Hamza, and
  Poiesi]{caraffa2024freeze}
Andrea Caraffa, Davide Boscaini, Amir Hamza, and Fabio Poiesi.
\newblock {FreeZe}: Training-free zero-shot {6D} pose estimation with geometric
  and vision foundation models.
\newblock In \emph{European Conference on Computer Vision (ECCV)}, 2024.

\bibitem[Chen et~al.(2023)Chen, Liu, Zhang, Qi, and Jia]{chen2023voxelnext}
Yukang Chen, Jianhui Liu, Xiangyu Zhang, Xiaojuan Qi, and Jiaya Jia.
\newblock {VoxelNeXt}: Fully sparse {VoxelNet} for {3D} object detection and
  tracking.
\newblock In \emph{Proceedings of the IEEE/CVF Conference on Computer Vision
  and Pattern Recognition (CVPR)}, 2023.

\bibitem[Choy et~al.(2020)Choy, Dong, and Koltun]{choy2020dgr}
Christopher Choy, Wei Dong, and Vladlen Koltun.
\newblock Deep global registration.
\newblock In \emph{Proceedings of the IEEE/CVF Conference on Computer Vision
  and Pattern Recognition (CVPR)}, 2020.

\bibitem[Curless and Levoy(1996)]{curless1996volumetric}
Brian Curless and Marc Levoy.
\newblock A volumetric method for building complex models from range images.
\newblock In \emph{SIGGRAPH}, 1996.

\bibitem[Drost et~al.(2017)Drost, Ulrich, Bergmann, Härtinger, and
  Steger]{mvitodd}
Bertram Drost, Markus Ulrich, Paul Bergmann, Philipp Härtinger, and Carsten
  Steger.
\newblock Introducing mvtec itodd — a dataset for 3d object recognition in
  industry.
\newblock In \emph{2017 IEEE International Conference on Computer Vision
  Workshops (ICCVW)}, pages 2200--2208, 2017.

\bibitem[Duffhauss et~al.(2022)Duffhauss, Demmler, and
  Neumann]{duffhauss2022mv6d}
Fabian Duffhauss, Tobias Demmler, and Gerhard Neumann.
\newblock {MV6D}: Multi-view {6D} pose estimation on {RGB-D} frames using a
  deep point-wise voting network.
\newblock In \emph{IEEE/RSJ International Conference on Intelligent Robots and
  Systems (IROS)}, 2022.

\bibitem[Duffhauss et~al.(2023)Duffhauss, Koch, Ziesche, Vien, and
  Neumann]{duffhauss2023symfm6d}
Fabian Duffhauss, Sebastian Koch, Hanna Ziesche, Ngo~Anh Vien, and Gerhard
  Neumann.
\newblock {SyMFM6D}: Symmetry-aware multi-directional fusion for multi-view
  {6D} object pose estimation.
\newblock \emph{IEEE Robotics and Automation Letters}, 8, 2023.

\bibitem[Ester et~al.(1996)Ester, Kriegel, Sander, and Xu]{ester1996dbscan}
Martin Ester, Hans-Peter Kriegel, J{\"o}rg Sander, and Xiaowei Xu.
\newblock A density-based algorithm for discovering clusters in large spatial
  databases with noise.
\newblock In \emph{International Conference on Knowledge Discovery and Data
  Mining (KDD)}, 1996.

\bibitem[Fan et~al.(2022)Fan, Pang, Zhang, Wang, Zhao, Wang, Wang, and
  Zhang]{fan2022sst}
Lue Fan, Ziqi Pang, Tianyuan Zhang, Yu-Xiong Wang, Hang Zhao, Feng Wang, Naiyan
  Wang, and Zhaoxiang Zhang.
\newblock Embracing single stride {3D} object detector with sparse transformer.
\newblock In \emph{Proceedings of the IEEE/CVF Conference on Computer Vision
  and Pattern Recognition (CVPR)}, 2022.

\bibitem[Gao et~al.(2021)Gao, Lauri, Hu, Zhang, and Frintrop]{gao2021cloudaae}
Ge Gao, Mikko Lauri, Xiaolin Hu, Jianwei Zhang, and Simone Frintrop.
\newblock {CloudAAE}: Learning {6D} object pose regression with on-line data
  synthesis on point clouds.
\newblock In \emph{IEEE International Conference on Robotics and Automation
  (ICRA)}, 2021.

\bibitem[Graham et~al.(2018)Graham, Engelcke, and van~der
  Maaten]{graham2018submanifold}
Benjamin Graham, Martin Engelcke, and Laurens van~der Maaten.
\newblock {3D} semantic segmentation with submanifold sparse convolutional
  networks.
\newblock In \emph{Proceedings of the IEEE/CVF Conference on Computer Vision
  and Pattern Recognition (CVPR)}, 2018.

\bibitem[He et~al.(2020)He, Sun, Huang, Liu, Fan, and Sun]{he2020pvn3d}
Yisheng He, Wei Sun, Haibin Huang, Jianran Liu, Haoqiang Fan, and Jian Sun.
\newblock {PVN3D}: A deep point-wise {3D} keypoints voting network for {6DoF}
  pose estimation.
\newblock In \emph{Proceedings of the IEEE/CVF Conference on Computer Vision
  and Pattern Recognition (CVPR)}, 2020.

\bibitem[He et~al.(2021)He, Huang, Fan, Chen, and Sun]{he2021ffb6d}
Yisheng He, Haibin Huang, Haoqiang Fan, Qifeng Chen, and Jian Sun.
\newblock {FFB6D}: A full flow bidirectional fusion network for {6D} pose
  estimation.
\newblock In \emph{Proceedings of the IEEE/CVF Conference on Computer Vision
  and Pattern Recognition (CVPR)}, 2021.

\bibitem[Hinton et~al.(2015)Hinton, Vinyals, and Dean]{hinton2015distilling}
Geoffrey Hinton, Oriol Vinyals, and Jeff Dean.
\newblock Distilling the knowledge in a neural network.
\newblock In \emph{NeurIPS Deep Learning and Representation Learning Workshop},
  2015.

\bibitem[Hodan et~al.(2018)Hodan, Michel, Brachmann, Kehl, Buch, Kraft, Drost,
  Vidal, Ihrke, Zabulis, Sahin, Manhardt, Tombari, Kim, Matas, and
  Rother]{hodan2018bop}
Tomas Hodan, Frank Michel, Eric Brachmann, Wadim Kehl, Anders~Glent Buch, Dirk
  Kraft, Bertram Drost, Joel Vidal, Stephan Ihrke, Xenophon Zabulis, Caner
  Sahin, Fabian Manhardt, Federico Tombari, Tae-Kyun Kim, Jiri Matas, and
  Carsten Rother.
\newblock {BOP}: Benchmark for {6D} object pose estimation.
\newblock In \emph{European Conference on Computer Vision (ECCV)}, 2018.

\bibitem[Horn(1987)]{horn1987closed}
Berthold K.~P. Horn.
\newblock Closed-form solution of absolute orientation using unit quaternions.
\newblock \emph{Journal of the Optical Society of America A}, 4\penalty0 (4),
  1987.

\bibitem[Hou et~al.(2022)Hou, Zhu, Ma, Loy, and Li]{hou2022pvkd}
Yuenan Hou, Xinge Zhu, Yuexin Ma, Chen~Change Loy, and Yikang Li.
\newblock Point-to-voxel knowledge distillation for {LiDAR} semantic
  segmentation.
\newblock In \emph{Proceedings of the IEEE/CVF Conference on Computer Vision
  and Pattern Recognition (CVPR)}, 2022.

\bibitem[Huang et~al.(2024)Huang, Yu, Yu, Navab, Ilic, and
  Busam]{huang2024matchu}
Junwen Huang, Hao Yu, Kuan-Ting Yu, Nassir Navab, Slobodan Ilic, and Benjamin
  Busam.
\newblock Matchu: Matching unseen objects for 6d pose estimation from rgb-d
  images.
\newblock In \emph{2024 IEEE/CVF Conference on Computer Vision and Pattern
  Recognition (CVPR)}, pages 10095--10105. IEEE, 2024.

\bibitem[Huang et~al.(2026)Huang, Hu, Yu, Ilic, Sundermeyer, and
  Busam]{xyz-ibd}
Junwen Huang, Jiaqi Hu, Peter~KT Yu, Slobodan Ilic, Martin Sundermeyer, and
  Benjamin Busam.
\newblock Xyz-ibd: Benchmarking robust 6d object pose estimation under
  real-world industrial complexity.
\newblock In \emph{European Conference on Computer Vision}, pages 376--393.
  Springer, 2026.

\bibitem[Kabsch(1976)]{kabsch1976}
Wolfgang Kabsch.
\newblock A solution for the best rotation to relate two sets of vectors.
\newblock \emph{Acta Crystallographica Section A}, 32\penalty0 (5), 1976.

\bibitem[Kalra et~al.(2024)Kalra, Stoppi, Marin, Taamazyan, Shandilya, Agarwal,
  Boykov, Chong, and Stark]{kalra2024ipd}
Agastya Kalra, Guy Stoppi, Dmitrii Marin, Vage Taamazyan, Aarrushi Shandilya,
  Rishav Agarwal, Anton Boykov, Tze Chong, and Michael Stark.
\newblock Towards co-evaluation of cameras, {HDR}, and algorithms for
  industrial-grade {6DoF} pose estimation.
\newblock In \emph{Proceedings of the IEEE/CVF Conference on Computer Vision
  and Pattern Recognition (CVPR)}, 2024.

\bibitem[Kalra et~al.(2026)Kalra, Salzmann, Stoppi, Marin, Agarwal, Taamazyan,
  Bokeloh, Hinterstoisser, Boykov, Dall'Olio, Dangol, Venkataraman, and
  Chen]{kalra2026threept}
Agastya Kalra, Tim Salzmann, Guy Stoppi, Dmitrii Marin, Rishav Agarwal, Vage
  Taamazyan, Martin Bokeloh, Stefan Hinterstoisser, Anton Boykov, Alberto
  Dall'Olio, Pravin Dangol, Kartik Venkataraman, and Huaijin Chen.
\newblock 3d-object perception transformer (3pt).
\newblock In \emph{Proceedings of the IEEE/CVF Conference on Computer Vision
  and Pattern Recognition (CVPR)}, pages 25777--25787, 2026.

\bibitem[Labb{\'e} et~al.(2020)Labb{\'e}, Carpentier, Aubry, and
  Sivic]{labbe2020cosypose}
Yann Labb{\'e}, Justin Carpentier, Mathieu Aubry, and Josef Sivic.
\newblock {CosyPose}: Consistent multi-view multi-object {6D} pose estimation.
\newblock In \emph{European Conference on Computer Vision (ECCV)}, 2020.

\bibitem[Labb{\'e} et~al.(2022)Labb{\'e}, Manuelli, Mousavian, Tyree,
  Birchfield, Tremblay, Carpentier, Aubry, Fox, and Sivic]{labbe2022megapose}
Yann Labb{\'e}, Lucas Manuelli, Arsalan Mousavian, Stephen Tyree, Stan
  Birchfield, Jonathan Tremblay, Justin Carpentier, Mathieu Aubry, Dieter Fox,
  and Josef Sivic.
\newblock {MegaPose}: {6D} pose estimation of novel objects via render \&
  compare.
\newblock In \emph{Conference on Robot Learning (CoRL)}, 2022.

\bibitem[Lai et~al.(2022)Lai, Liu, Jiang, Wang, Zhao, Liu, Qi, and
  Jia]{lai2022stratified}
Xin Lai, Jianhui Liu, Li Jiang, Liwei Wang, Hengshuang Zhao, Shu Liu, Xiaojuan
  Qi, and Jiaya Jia.
\newblock Stratified transformer for {3D} point cloud segmentation.
\newblock In \emph{Proceedings of the IEEE/CVF Conference on Computer Vision
  and Pattern Recognition (CVPR)}, 2022.

\bibitem[Lai et~al.(2023)Lai, Chen, Lu, Liu, and Jia]{lai2023sphereformer}
Xin Lai, Yukang Chen, Fanbin Lu, Jianhui Liu, and Jiaya Jia.
\newblock Spherical transformer for {LiDAR}-based {3D} recognition.
\newblock In \emph{Proceedings of the IEEE/CVF Conference on Computer Vision
  and Pattern Recognition (CVPR)}, 2023.

\bibitem[Law and Deng(2018)]{law2018cornernet}
Hei Law and Jia Deng.
\newblock {CornerNet}: Detecting objects as paired keypoints.
\newblock In \emph{European Conference on Computer Vision (ECCV)}, 2018.

\bibitem[Leuze et~al.(2026)Leuze, Hoh, Do{\u{g}}an, R.-Pe{\~n}a, and
  Schoettl]{leuze2026sdt6d}
Nico Leuze, Maximilian Hoh, Samed Do{\u{g}}an, Nicolas R.-Pe{\~n}a, and Alfred
  Schoettl.
\newblock {SDT-6D}: Fully sparse depth-transformer for staged end-to-end {6D}
  pose estimation in industrial multi-view bin picking.
\newblock In \emph{Proceedings of the IEEE/CVF Winter Conference on
  Applications of Computer Vision (WACV)}, 2026.

\bibitem[Li and Schoellig(2023)]{li2023mvkp}
Alan Li and Angela~P. Schoellig.
\newblock Multi-view keypoints for reliable {6D} object pose estimation.
\newblock In \emph{IEEE International Conference on Robotics and Automation
  (ICRA)}, 2023.

\bibitem[Li and Stamos(2023)]{li2023swindepose}
Zhujun Li and Ioannis Stamos.
\newblock Depth-based {6DoF} object pose estimation using swin transformer.
\newblock In \emph{IEEE/RSJ International Conference on Intelligent Robots and
  Systems (IROS)}, 2023.

\bibitem[Lin et~al.(2024{\natexlab{a}})Lin, Liu, Lu, and Jia]{lin2024sam6d}
Jiehong Lin, Lihua Liu, Dekun Lu, and Kui Jia.
\newblock {SAM-6D}: Segment anything model meets zero-shot {6D} object pose
  estimation.
\newblock In \emph{Proceedings of the IEEE/CVF Conference on Computer Vision
  and Pattern Recognition (CVPR)}, 2024{\natexlab{a}}.

\bibitem[Lin et~al.(2017)Lin, Goyal, Girshick, He, and
  Doll{\'a}r]{lin2017focal}
Tsung-Yi Lin, Priya Goyal, Ross Girshick, Kaiming He, and Piotr Doll{\'a}r.
\newblock Focal loss for dense object detection.
\newblock In \emph{Proceedings of the IEEE/CVF International Conference on
  Computer Vision (ICCV)}, 2017.

\bibitem[Lin et~al.(2024{\natexlab{b}})Lin, Wang, Zhou, Liu, and
  Chen]{lin2024transpose}
Xiao Lin, Deming Wang, Guangliang Zhou, Chengju Liu, and Qijun Chen.
\newblock {TransPose}: {6D} object pose estimation with geometry-aware
  transformer.
\newblock \emph{Neurocomputing}, 589, 2024{\natexlab{b}}.

\bibitem[Mike{\v{s}}t{\'\i}kov{\'a} et~al.(2026)Mike{\v{s}}t{\'\i}kov{\'a},
  Fourmy, C{\'\i}fka, Sivic, and Petrik]{alignpose2025}
Anna~{\v{S}}{\'a}rov{\'a} Mike{\v{s}}t{\'\i}kov{\'a}, M{\'e}d{\'e}ric Fourmy,
  Martin C{\'\i}fka, Josef Sivic, and Vladimir Petrik.
\newblock Alignpose: Generalizable 6d pose estimation via multi-view
  feature-metric alignment.
\newblock In \emph{Proceedings of the IEEE/CVF Conference on Computer Vision
  and Pattern Recognition}, pages 14626--14636, 2026.

\bibitem[Newcombe et~al.(2011)Newcombe, Izadi, Hilliges, Molyneaux, Kim,
  Davison, Kohli, Shotton, Hodges, and Fitzgibbon]{newcombe2011kinectfusion}
Richard~A. Newcombe, Shahram Izadi, Otmar Hilliges, David Molyneaux, David Kim,
  Andrew~J. Davison, Pushmeet Kohli, Jamie Shotton, Steve Hodges, and Andrew
  Fitzgibbon.
\newblock {KinectFusion}: Real-time dense surface mapping and tracking.
\newblock In \emph{IEEE International Symposium on Mixed and Augmented Reality
  (ISMAR)}, 2011.

\bibitem[Nguyen et~al.(2025)Nguyen, Tyree, Guo, Fourmy, Gouda, Lee, Moon, Son,
  Ranftl, Tremblay, Brachmann, Drost, Lepetit, Rother, Birchfield, Matas,
  Labbe, Sundermeyer, and Hodan]{nguyen2025bop}
Van~Nguyen Nguyen, Stephen Tyree, Andrew Guo, Mederic Fourmy, Anas Gouda,
  Taeyeop Lee, Sungphill Moon, Hyeontae Son, Lukas Ranftl, Jonathan Tremblay,
  Eric Brachmann, Bertram Drost, Vincent Lepetit, Carsten Rother, Stan
  Birchfield, Jiri Matas, Yann Labbe, Martin Sundermeyer, and Tomas Hodan.
\newblock {BOP} challenge 2024 on model-based and model-free {6D} object pose
  estimation.
\newblock In \emph{IEEE/CVF Conference on Computer Vision and Pattern
  Recognition Workshops (CVPRW)}, 2025.

\bibitem[Park et~al.(2019)Park, Patten, and Vincze]{park2019pix2pose}
Kiru Park, Timothy Patten, and Markus Vincze.
\newblock {Pix2Pose}: Pixel-wise coordinate regression of objects for {6D} pose
  estimation.
\newblock In \emph{Proceedings of the IEEE/CVF International Conference on
  Computer Vision (ICCV)}, 2019.

\bibitem[Perez et~al.(2018)Perez, Strub, de~Vries, Dumoulin, and
  Courville]{perez2018film}
Ethan Perez, Florian Strub, Harm de Vries, Vincent Dumoulin, and Aaron
  Courville.
\newblock {FiLM}: Visual reasoning with a general conditioning layer.
\newblock In \emph{AAAI Conference on Artificial Intelligence}, 2018.

\bibitem[Qi et~al.(2017)Qi, Su, Mo, and Guibas]{qi2017pointnet}
Charles~R. Qi, Hao Su, Kaichun Mo, and Leonidas~J. Guibas.
\newblock {PointNet}: Deep learning on point sets for {3D} classification and
  segmentation.
\newblock In \emph{Proceedings of the IEEE/CVF Conference on Computer Vision
  and Pattern Recognition (CVPR)}, 2017.

\bibitem[Schaub et~al.(2024)Schaub, Wolff, Hoh, and Schöttl]{henry_iros}
Henry Schaub, Christian Wolff, Maximilian Hoh, and Alfred Schöttl.
\newblock Probabilistic closed-loop active grasping.
\newblock In \emph{IEEE/RSJ International Conference on Intelligent Robots and
  Systems (IROS)}, 2024.

\bibitem[Shi et~al.(2021)Shi, Huang, Xu, Zhang, and Xu]{shi2021stablepose}
Yifei Shi, Junwen Huang, Xin Xu, Yifan Zhang, and Kai Xu.
\newblock {StablePose}: Learning {6D} object poses from geometrically stable
  patches.
\newblock In \emph{Proceedings of the IEEE/CVF Conference on Computer Vision
  and Pattern Recognition (CVPR)}, 2021.

\bibitem[Wang et~al.(2021)Wang, Manhardt, Tombari, and Ji]{wang2021gdrnet}
Gu Wang, Fabian Manhardt, Federico Tombari, and Xiangyang Ji.
\newblock {GDR-Net}: Geometry-guided direct regression network for monocular
  {6D} object pose estimation.
\newblock In \emph{Proceedings of the IEEE/CVF Conference on Computer Vision
  and Pattern Recognition (CVPR)}, 2021.

\bibitem[Wang et~al.(2019)Wang, Sridhar, Huang, Valentin, Song, and
  Guibas]{wang2019nocs}
He Wang, Srinath Sridhar, Jingwei Huang, Julien Valentin, Shuran Song, and
  Leonidas~J. Guibas.
\newblock Normalized object coordinate space for category-level {6D} object
  pose and size estimation.
\newblock In \emph{Proceedings of the IEEE/CVF Conference on Computer Vision
  and Pattern Recognition (CVPR)}, 2019.

\bibitem[Wang et~al.(2025)Wang, Hu, Zhuo, and Luo]{wang2025frtpose}
Yulin Wang, Mengting Hu, Jianghao Zhuo, and Chen Luo.
\newblock {FRT-Pose WAPRv2} (multi-cam).
\newblock \url{https://bop.felk.cvut.cz/method_info/1157/}, 2025.
\newblock Accessed: 2026-08-18.

\bibitem[Wen et~al.(2024)Wen, Yang, Kautz, and
  Birchfield]{wen2024foundationpose}
Bowen Wen, Wei Yang, Jan Kautz, and Stan Birchfield.
\newblock {FoundationPose}: Unified {6D} pose estimation and tracking of novel
  objects.
\newblock In \emph{Proceedings of the IEEE/CVF Conference on Computer Vision
  and Pattern Recognition (CVPR)}, 2024.

\bibitem[Wu et~al.(2022)Wu, Lao, Jiang, Liu, and Zhao]{ptv2}
Xiaoyang Wu, Yixing Lao, Li Jiang, Xihui Liu, and Hengshuang Zhao.
\newblock Point transformer v2: Grouped vector attention and partition-based
  pooling.
\newblock In \emph{NeurIPS}, 2022.

\bibitem[Wu et~al.(2024)Wu, Jiang, Wang, Liu, Liu, Qiao, Ouyang, He, and
  Zhao]{wu2024ptv3}
Xiaoyang Wu, Li Jiang, Peng-Shuai Wang, Zhijian Liu, Xihui Liu, Yu Qiao, Wanli
  Ouyang, Tong He, and Hengshuang Zhao.
\newblock Point transformer {V3}: Simpler, faster, stronger.
\newblock In \emph{Proceedings of the IEEE/CVF Conference on Computer Vision
  and Pattern Recognition (CVPR)}, 2024.

\bibitem[Yang et~al.(2021)Yang, Gao, Li, and Waslander]{yang2021robi}
Jun Yang, Yizhou Gao, Dong Li, and Steven~L. Waslander.
\newblock {ROBI}: A multi-view dataset for reflective objects in robotic
  bin-picking.
\newblock In \emph{IEEE/RSJ International Conference on Intelligent Robots and
  Systems (IROS)}, 2021.

\bibitem[Yang et~al.(2022)Yang, Shi, Ding, Wang, and Qi]{yang2022towards}
Jihan Yang, Shaoshuai Shi, Runyu Ding, Zhe Wang, and Xiaojuan Qi.
\newblock Towards efficient {3D} object detection with knowledge distillation.
\newblock In \emph{Advances in Neural Information Processing Systems
  (NeurIPS)}, 2022.

\bibitem[Yang et~al.(2025)Yang, Xue, Ghavidel, and Waslander]{yang2025active}
Jun Yang, Wenjie Xue, Sahar Ghavidel, and Steven~L. Waslander.
\newblock Active {6D} pose estimation for textureless objects using multi-view
  {RGB} frames.
\newblock \emph{The International Journal of Robotics Research}, 2025.

\bibitem[Zhang et~al.(2023)Zhang, Dong, Tai, and Ma]{zhang2023pointdistiller}
Linfeng Zhang, Runpei Dong, Hung-Shuo Tai, and Kaisheng Ma.
\newblock {PointDistiller}: Structured knowledge distillation towards efficient
  and compact {3D} detection.
\newblock In \emph{Proceedings of the IEEE/CVF Conference on Computer Vision
  and Pattern Recognition (CVPR)}, 2023.

\bibitem[Zhou et~al.(2018)Zhou, Park, and Koltun]{zhou2018open3d}
Qian-Yi Zhou, Jaesik Park, and Vladlen Koltun.
\newblock {Open3D}: A modern library for {3D} data processing.
\newblock \emph{arXiv:1801.09847}, 2018.

\bibitem[Zhou et~al.(2019)Zhou, Barnes, Lu, Yang, and Li]{zhou2019continuity}
Yi Zhou, Connelly Barnes, Jingwan Lu, Jimei Yang, and Hao Li.
\newblock On the continuity of rotation representations in neural networks.
\newblock In \emph{Proceedings of the IEEE/CVF Conference on Computer Vision
  and Pattern Recognition (CVPR)}, 2019.

\end{thebibliography}
